%% file: sample-sigconf.tex
\documentclass[sigconf]{acmart}
\AtBeginDocument{%
  }

\setcopyright{cc}
\setcctype[4.0]{by-nc-sa}
\usepackage{booktabs}
\usepackage{multirow}
\usepackage{xcolor}
\usepackage{xspace}
\usepackage{graphicx}
\usepackage{amsmath}
\usepackage{subfigure}
\newcommand{\fd}{FraudSquad\xspace}
\newcommand{\llms}{LLMs\xspace}
\newcommand{\llamathree}{Llama3\xspace}
\newcommand{\qwen}{Qwen2\xspace}
\newcommand{\ds}{Qwen-DSR1\xspace}
\newcommand{\gpto}{GPT-4.1\xspace}
\newcommand{\dyelpres}{\texttt{Yelp}\xspace}
\newcommand{\dama}{\texttt{Amazon}\xspace}
\newcommand{\dqa}{\texttt{ChineseQA}\xspace}
\newcommand{\para}[1]{\smallskip\noindent\textbf{#1}}
 \newcommand{\todo}[1]{\xspace}

\definecolor{myorgance}{RGB}{246, 201, 193}
\definecolor{myyellow}{RGB}{251, 234, 196}
\definecolor{myblue}{RGB}{198, 221, 252}
\definecolor{myblue2}{RGB}{66, 121, 186}
\definecolor{mygreen}{RGB}{165, 209, 220}
\definecolor{mypurple}{RGB}{215, 188, 237}
\newcommand{\patterna}[1]{#1}
\newcommand{\patternb}[1]{#1}
\newcommand{\patternc}[1]{#1}
\newcommand{\patternd}[1]{#1}
\newcommand{\patterne}[1]{#1}

\begin{document}

\title{Detecting LLM-Generated Spam Reviews by Integrating Language Model Embeddings and Graph Neural Network}


\author{Xin Liu}
\authornote{Both authors contributed equally to this research.}
\email{liuxin19@tsinghua.org.cn}
\affiliation{%
  \institution{Tsinghua University}
  \city{Beijing}
  \country{China}
}
\email{liuxin4@supcon.com}
\affiliation{%
  \institution{SUPCON}
  \city{Hangzhou}
  \country{China}
}

\author{Rongwu Xu}
\authornotemark[1]
\email{xrw22@mails.tsinghua.edu.cn}
\affiliation{%
  \institution{Tsinghua University}
\city{Beijing}
  \country{China}
}

\author{Xinyi Jia}
\email{jiaxy21@mails.tsinghua.edu.cn}
\affiliation{%
  \institution{Tsinghua University}
\city{Beijing}
  \country{China}
}

\author{Jason Liao}
\email{jliao8@student.ubc.ca}
\affiliation{%
 \institution{University of British Columbia}
 \city{Vancouver}
 \state{British Columbia}
 \country{Canada}}

\author{Jiao Sun}
\email{jiaosun.thu@gmail.com}
\affiliation{%
  \institution{Google DeepMind}
  \city{Mountain View}
  \state{California}
  \country{USA}}

\author{Ling Huang}
\authornote{Both authors are corresponding authors.}
\email{linghuang@fintec.ai}
\affiliation{%
  \institution{AHI Fintech}
  \city{Beijing}
  \country{China}}

\author{Wei Xu}
\authornotemark[2]
\email{weixu@tsinghua.edu.cn}
\affiliation{%
  \institution{Tsinghua University}
  \city{Beijing}
  \country{China}}


\begin{abstract}
The rise of large language models (LLMs) has enabled the generation of highly persuasive spam reviews that closely mimic human writing. These reviews pose significant challenges for existing detection systems and threaten the credibility of online platforms. In this work, we first create \emph{three realistic LLM-generated spam review datasets} using three distinct LLMs, each guided by product metadata and genuine reference reviews. Evaluations by GPT-4.1 confirm the high persuasion and deceptive potential of these reviews.

To address this threat, we propose \textbf{\fd}, \emph{a hybrid detection model} that integrates text embeddings from a pre-trained language model with a gated graph transformer for spam node classification. \fd captures both semantic and behavioral signals without relying on complex feature engineering or massive training resources.
Experiments show that \fd outperforms state-of-the-art baselines by up to 44.22\% in precision and 43.01\% in recall on three LLM-generated datasets, while also achieving promising results on two human-written spam datasets.
Furthermore, \fd maintains a modest model size and requires minimal labeled training data, making it a practical solution for real-world applications. 
Our contributions include new synthetic datasets, a practical detection framework, and empirical evidence highlighting the urgency of adapting spam detection to the LLM era. Our code and datasets are available at: \url{https://anonymous.4open.science/r/FraudSquad-5389/}.
\end{abstract}





\maketitle

\input{1introduction.tex}

\input{2method.tex}

\input{3experiment_sim.tex}

\input{4related_work.tex}

\input{5conclusion.tex}

\input{6limitation.tex}
\input{7appendix.tex}

\clearpage

\section*{GenAI Usage Disclosure}
In this work, generative AI (GenAI) tools were primarily used to create the three datasets of LLM-generated spam reviews, as already described in Section~\ref{sec:syn_data}. These datasets were synthesized through a structured generation pipeline utilizing large language models.

Part of the experimental codebase was adapted from publicly available implementations~\cite{Xiang2023semi}, while the remaining components were developed independently. We used AI-assisted tools (Cursor more specifically) mainly to implement the computation of average pair-wise BLEU scores in review text evaluation.

The manuscript’s core content was written by the authors. AI tools (GPT-4o) were used solely for minor editing and language refinement to improve clarity and fluency.

\bibliographystyle{ACM-Reference-Format}
\bibliography{combined_ref}
\end{document}

%% file: 1introduction.tex
\section{Introduction}

\begin{figure}[h]
    \centering
    \includegraphics[width=0.95\linewidth]{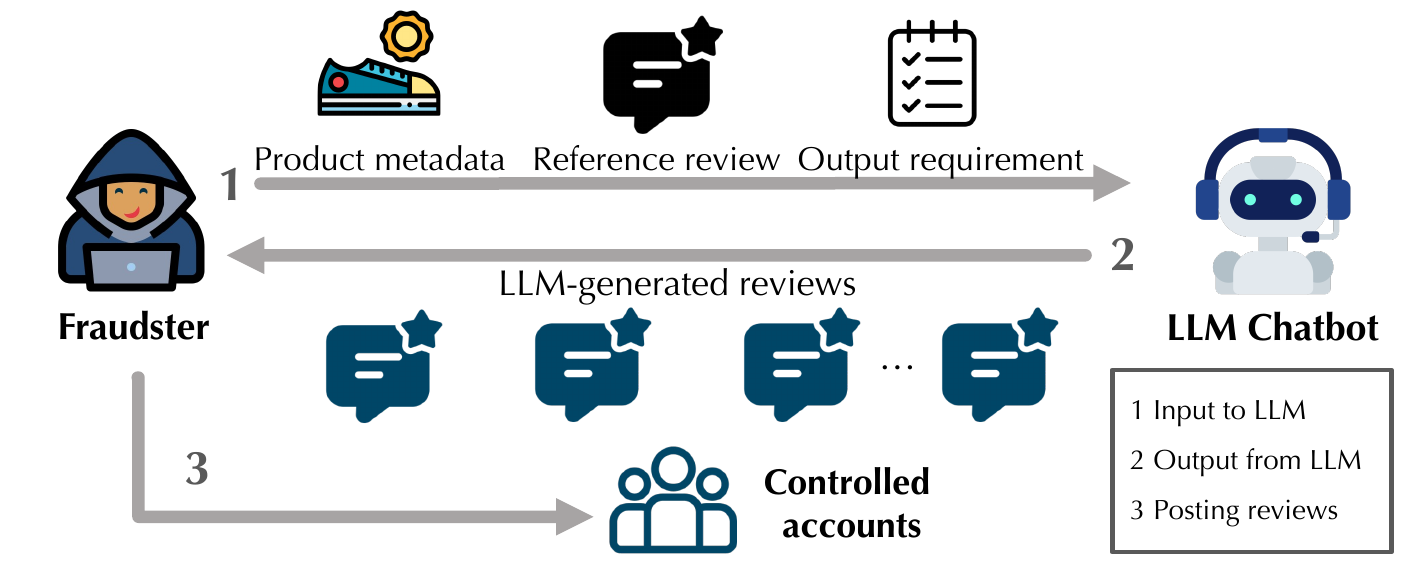}
    \caption{Workflow of LLM-generated review spamming. There are three steps. Step 1: Provide the LLM-based chatbot with the product metadata, genuine reference review texts, and specific output requirements. Step 2: Obtain the review texts generated by the LLM based on the input. Step 3: Use controlled accounts to post the generated reviews.}
    \label{fig:spam_workflow}
\end{figure}

Online reviews play a pivotal role in shaping consumer decision-making and influencing business reputations \cite{Duma2023ADH}. However, the proliferation of spam reviews, i.e., deceptive content designed to mislead consumers, has emerged as a significant challenge \cite{Jindal2008,Andresini2022ReviewSD}.
Fraudsters often \emph{control multiple accounts to post coordinated spam reviews}, manipulating a target's reputation for profit \cite{Xu2013CopyCatch, caoSynchroTrap}.
These fake reviews not only distort consumer trust but also erode the credibility of platforms like Amazon and Yelp. The economic impact is substantial, with spam reviews estimated to cost consumers and businesses around \$152 billion annually \cite{economic_toll}. Detecting and curbing spam reviews is therefore essential for protecting consumer interests and ensuring the integrity of online review ecosystems.

However, detecting spam reviews remains challenging in two key aspects. \textbf{\emph{First}}, recent advances in large language models (LLMs) \cite{zhao2023survey}, such as ChatGPT \cite{openai2023gpt4}, Llama \cite{touvron2023llama, touvron2023llama2, grattafiori2024llama3herdmodels}, and DeepSeek \cite{deepseekai2025}, have made it easier to generate sophisticated, deceptive content \cite{Wu2024Fake,feng2024does}, heightening the urgency of effective detection.
In particular, fraudsters can now exploit publicly available information to craft realistic fake reviews using LLMs \cite{xu2025nucleardeployedanalyzingcatastrophic}. As illustrated in Figure \ref{fig:spam_workflow}, a fraudster may input product metadata, genuine reference reviews, and specific output instructions into an LLM-based chatbot. The generated reviews are then posted through controlled accounts, making them appear authentic and difficult to detect.

Following the fraudster's workflow illustrated in Figure \ref{fig:spam_workflow}, we first constructed three LLM-generated spam review datasets to fill the lack of publicly available data. The generation pipeline was applied to three distinct LLMs using real-world review data from Amazon.  Each LLM was prompted with product metadata, reference review texts, and specific output requirements to simulate realistic spam generation scenarios and generated 2,500  spam reviews.
We employed \gpto to evaluate the quality of LLM-generated reviews. The assessment revealed that these reviews are \emph{highly persuasive to potential consumers and closely resemble genuine human-written content}. Notably, the evaluation scores of the generated reviews surpassed those of genuine human-written reviews in being persuasive, detailed, convincing and influential on a five-point Likert scale. Furthermore, the generated reviews well met the output requirements specified by the fraudster, underscoring the urgency for developing accurate detection methods to mitigate the potential impact of such sophisticated spam content.

\textbf{\emph{Second}}, existing spam/fraud detection methods often overlook the rich linguistic features embedded in review texts, which can provide crucial insights for identifying sophisticated spam reviews generated from LLMs. While graph-based approaches \cite{tianCFD, hooi2016Fraudar}, including recent advancements in Graph Neural Networks (GNNs) \cite{Zhang2020GCN}, have demonstrated effectiveness in capturing complex interactions within review graphs, they predominantly rely on \emph{engineered features} derived from review contents to distinguish spamming reviews from fraudsters and non-spamming reviews from genuine users. This dependence limits their capacity to detect nuanced LLM-generated spam reviews that closely mimic authentic human writing.

To address the new challenge of detecting LLM-generated spam reviews, we propose a novel hybrid detection framework, namely \textbf{\fd}, which integrates language model-enhanced node embeddings with graph neural networks. \fd introduces two key innovations: first, it enriches node representations using text embeddings from a pre-trained language model; second, it employs gated graph transformers to capture relationships among review nodes within a constructed review graph for spam classification. This design allows \fd to leverage both the linguistic content of review texts and user behavioral patterns—specifically, actions such as rating a product at a given time. By carefully selecting a lightweight language model for text embeddings, \fd remains both efficient and effective, achieving high detection accuracy without relying on complex feature engineering \cite{liu2021pick, Xiang2023semi, dgagnn} or extensive training resources. 

Experimental results show that our method \fd can accurately detect these LLM-generated spam reviews, outperforming state-of-the-art fraud detectors \cite{liu2021pick, Xiang2023semi, dgagnn} by up to 44.22\% in precision and 43.01\% in recall. \fd achieves overall metric scores of 89.45\%-99.98\% with only 1\% annotated labels at training time. In addition, we find \fd is also significantly more effective on two human-written spam review datasets. The ablation studies verify that advanced text embedding and graph structure are indispensable for accurate detection, saving the labor of maintaining engineered features without requiring massive training resources. 

On the whole, our contributions are threefold:
\begin{enumerate}
    \item We are the first to comprehensively study the \textbf{\emph{problem}} of detecting LLM-generated spam reviews.
    \item We synthesize three realistic \textbf{\emph{datasets}} of LLM-generated spam reviews and evaluate the quality of the generated texts from multiple perspectives.
    \item We propose a detection \textbf{\emph{model}} \fd that integrates language model embeddings and gated graph transformers, which achieves state-of-the-art detection performance without feature engineering or massive training resources.
\end{enumerate}

%% file: 2method.tex
\section{Synthesizing LLM-Generated Spam Review Datasets}\label{sec:syn_data}

\begin{table}[tb]
\centering
\caption{Prompt for generating the review texts using \llms. Inputs related to product metadata, reference review texts, and output requirements are included in \textcolor{myblue2}{[]}.}
\label{table:gene_prompt}
\fontsize{8}{8}\selectfont
\begin{tabular}{p{8cm}}
\toprule
I need your help to write reviews for a product \textcolor{myblue2}{[product name]} on Amazon in the category of \textcolor{myblue2}{[product category]}. 
The official description of the product given by the store is as follows:  
\textcolor{myblue2}{[product official description]} \\
Besides, I will give you a set of review of this product for reference: 
\textcolor{myblue2}{[reference review texts]} \\
Now, please output \textcolor{myblue2}{[review number]} \textcolor{myblue2}{[positive/negative]} reviews. Each review contains no more than \textcolor{myblue2}{[max word]} words. Please write diversified reviews as if they were written by different customers, for example, with different lengths and styles. Start with another paragraph for each review and begin with Review 1. 2. 3., etc. \\\bottomrule
\end{tabular}
\end{table}

Since there are no public datasets specifically designed to test the detection of LLM-generated spam reviews, we create three datasets in this work. Each dataset contains the spam reviews generated from a distinct LLM.
Basically, we simulate a \textbf{\emph{fraudster}} by generating spam review texts using an \textbf{\emph{LLM}} and letting \textbf{\emph{controlled accounts}} post these reviews, as shown in Figure \ref{fig:spam_workflow}. It is important to note that we do not consider all LLM-generated reviews as spam by default. Instead, what we define as spam is the \emph{coordinated and repeated posting of such reviews by fake or controlled accounts across multiple products, typically with the intent of manipulation or profit}. In this process, LLMs are one \emph{component} used to facilitate spamming.

\para{Notation of review.} We consider a review as a specific tuple that contains the user, review text, rating star, product, and timestamp.

\begin{table*}[tb]
\centering
\caption{Performance statistics of spam review text generation using LLMs demonstrate that the output requirements are well met. (1) The LLMs closely follow the specified output format, producing the required number of reviews with high accuracy. (2) The length of the generated reviews is consistently close to the maximum limit of 100 words. (3) The average pairwise BLEU scores among reviews generated for the same product are low, indicating a high degree of diversity in content and style.} 
\label{table:generation-stats}
\begin{tabular}{@{}lrrrr@{}}
\toprule
LLM & Outputted/required & Max. \#words & Avg. \#words & Avg. pairwise BLEU \\ \midrule
\llamathree    &2488/2500   & 94 & 56.5 $\pm$ 7.4 & 0.05 $\pm$ 0.03\\
\qwen  &2500/2500     & 133  & 54.5 $\pm$ 8.7 & 0.03 $\pm$ 0.02  \\ 
\ds & 2500/2500 & 102 & 60.5 $\pm$ 10.0 & 0.10 $\pm$ 0.04 \\
\bottomrule
\end{tabular}
\end{table*}

\subsection{Generating Spam Review Texts}\label{sec:fake_case_study}

\para{Task description and the generation pipeline.} 
Suppose a fraudster wants to post spam reviews for a specific product. The goal is to generate \emph{review texts} that are both highly relevant to the target product and hard to distinguish from human-written ones.
To this end, the fraudster could use an LLM-based chat assistant, providing it with detailed information about the product. This may include product metadata and reference review texts, both of which can easily be collected from e-commerce platforms like Amazon.
In this work, we simulate such behavior by assuming the fraudster provides the following inputs to the LLM:

\begin{itemize}
	\item \emph{Product metadata:} Includes the product's name, category, and the official description provided by the seller.
	\item \emph{Reference review texts:} A set of genuine user review texts the product has received, which may be either positive or negative in sentiment.
	\item \emph{Output requirements:} Specifies the desired sentiment, output number, maximum word count, diversity in content length and style, and formatting requirements (e.g., each review should appear as a separate paragraph).
\end{itemize}

The generation pipeline follows the prompt in Table \ref{table:gene_prompt}. 
The fraudster sends a \texttt{user} message requesting to write reviews and providing the inputs marked within \textcolor{myblue2}{[]}. The fraudster receives the \texttt{assistant} message from LLM with the generated spam review texts.

\para{Generation setups.} 
We apply this generation pipeline to the \dama dataset, which is built upon the large-scale Amazon Review Dataset \cite{hou2024bridging}. Specifically, we select reviews from the year 2022 across eight product categories: Baby Products, Video Games, Software, Musical Instruments, Appliances, All Beauty, Health \& Personal Care, and Digital Music. In total, the derived \dama dataset contains 7,617 products and 86,758 reviews.

We focus on generating \emph{positive review texts} for \emph{low-performing products}, i.e., those with the lowest average star ratings and the fewest reviews. Since over 75\% of products in the \dama dataset have an average rating above 4.3 on a five-star scale, it is reasonable to assume that products falling below this threshold may seek to improve their reputation. We randomly select 500 such products and generate five positive reviews for each, with a maximum length of 100 words per review. For context, each generation is guided by the first genuine review of the target product, used as a reference.

\para{LLM selection.} 
Three \llms are leveraged for spam review text generation: (1) Qwen2-72B-Instruct \cite{qwen2} (\qwen); (2) Llama3-8B-Instruct\footnote{\url{https://github.com/meta-llama/llama3}} (\llamathree); and (3) DeepSeek-R1-Distill-Qwen-32B (\ds)\footnote{\url{https://huggingface.co/deepseek-ai/DeepSeek-R1-Distill-Qwen-32B}}. 
All models are open-source and implemented using the Ollama\footnote{\url{https://github.com/ollama/ollama}} framework. 
This process results in three separate datasets of LLM-generated spam review \emph{texts}, which we later use to build the three final spam review datasets (see Section \ref{subsec:final} for details).

\subsection{Evaluating LLM-Generated Spam Review Texts}

\begin{figure}[tb]
\centering
\includegraphics[width=8cm]{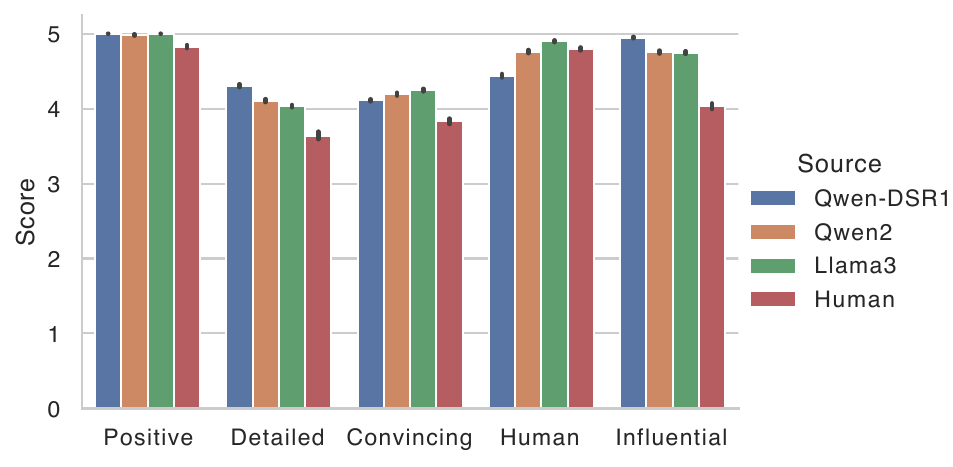}
\caption{Evaluation results of LLM-generated and human-written review texts by \gpto, rated on a Likert scale. The results show that LLM-generated reviews outperform human-written ones in terms of being positive, detailed, convincing, and influential. Moreover, they are often perceived as highly human-like in style and tone.} \label{fig:eval_llm_review}
\end{figure}

\para{Persuasiveness and human-likeness.}
First, we aim to assess whether \llms can generate spam review texts that are \emph{highly persuasive to potential customers} and \emph{closely resemble those written by real users}. To this end, we employ the advanced \gpto \cite{openai2025gpt41} model as an automatic evaluator \cite{gu2024survey} to rate the LLM-generated reviews across \textbf{\emph{five distinct dimensions}}: (1) whether the review is clearly \emph{positive}; (2) whether it is \emph{detailed}; (3) whether it is \emph{convincing}; (4) whether it appears to be written by a typical \emph{human}; and (5) whether it is \emph{influential} from a potential customer’s perspective. 
Each review is rated on a Likert scale from 0 to 5, where a higher score indicates better performance on the respective criterion. The full evaluation prompt is shown in Table \ref{table:eval_prompt}. 
For comparison, we also randomly sample 2,500 five-star human-written reviews from the \dama dataset and evaluate them using the same procedure.
Figure \ref{fig:eval_llm_review} shows the evaluation results from \gpto. All LLMs achieve consistently high scores (above 4) across all evaluation dimensions, with particularly strong performance in the positive and influential aspects.
Notably, LLM-generated reviews outperform human-written reviews in all areas except for the human-like aspect, which shows slightly diverged results. Two of the three LLMs perform on par with or better than human-written reviews, while \ds scores slightly lower. Overall, these results suggest that \emph{LLM-generated reviews can easily be mistaken for genuine human-written ones and may significantly influence potential customers, potentially by spreading misleading information.}

\para{Fulfilling output requirements.}
Second, we aim to evaluate whether the LLM-generated review texts meet the output requirements specified in the fraudster's prompts. Table \ref{table:generation-stats} summarizes the generation statistics. The number of outputted reviews is extremely close to the required number, with \llamathree and \ds matching it exactly. This indicates that all \llms follow the output format very reliably, enabling automatic extraction of review texts with a high success rate (over 99.5\%).
Besides, the word limit requirement is also well satisfied. In particular, \llamathree consistently produces reviews with fewer than 100 words, adhering strictly to the maximum length constraint.
To assess the diversity of reviews generated for the same product, we calculate the average pairwise BLEU scores \cite{papineni-etal-2002-bleu}. Lower scores indicate greater diversity. As shown in Table \ref{table:generation-stats}, all LLMs achieve low BLEU scores, suggesting that their outputs are varied and not repetitive. Additionally, a \emph{manual check} of 100 randomly selected reviews confirms that the texts differ significantly in wording and highlight various aspects of the product. For qualitative examples, refer to Appendix \ref{appen:examples}.

\subsection{Creating the Final LLM-Generated Spam Review Datasets}
\label{subsec:final}

To build the three final spam review datasets, \emph{not just the review texts}, we simulate how fraudsters might inject fake reviews into a real review platform. Specifically, we assume that fraudsters take over some existing user accounts from the original \dama dataset. These accounts had previously posted normal reviews, but are now used to post the LLM-generated spam reviews, each user posting only 2 generated spam reviews. This setup creates a more realistic and challenging detection scenario \cite{Dou2020EnhancingGN}.

We select compromised users based on how active they are—the more reviews a user has written, the more likely they are to be chosen. Each compromised user posts two fake five-star reviews for a target product. The posting time for these reviews is randomly chosen within five days of the product’s first real review, at any hour of the day.

In this way, we create three LLM-generated spam review datasets: \texttt{Amazon-\llamathree}, \texttt{Amazon-\qwen}, and \texttt{Amazon-\ds}. All three datasets contain the same genuine reviews in the \dama dataset but different LLM-generated spam reviews.

\begin{figure*}[ht]
    \centering
    \includegraphics[width=0.9\linewidth]{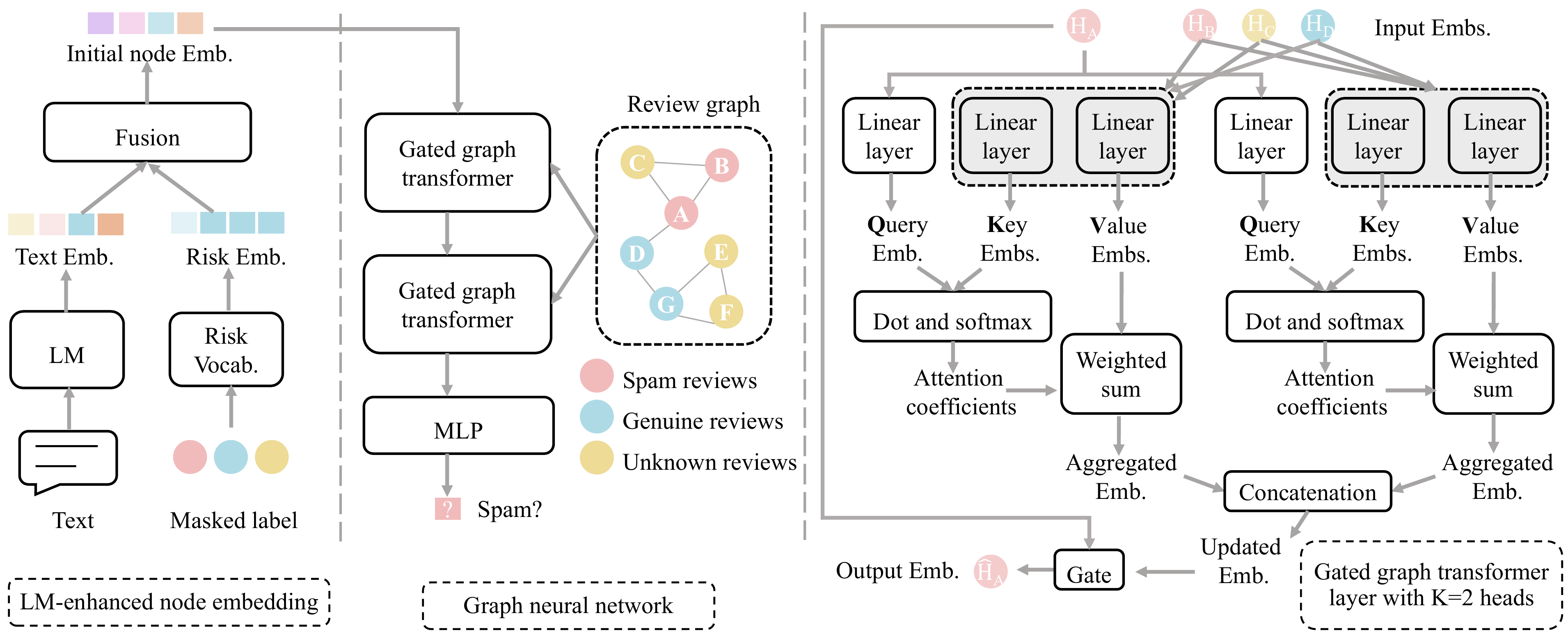}
    \caption{The overall architecture of \fd by integrating LM-enhanced node embedding and graph neural network on the constructed review graph, where the review nodes are connected by various relations. The graph neural network in \fd employs multi-headed gated graph transformer layers, which are also illustrated in this figure.}
    \label{fig:fraudsquad_architecture}
\end{figure*}

\section{\fd: A Novel Approach for Spam Review Detection}
In this section, we present a novel approach \fd for spam review detection, including the problem formulation and the model architecture in detail.

\subsection{Problem Formulation}

To detect spam reviews by considering \emph{both the review texts and fraudsters' behavior}, we frame the problem as a node classification task on a review graph $\mathcal G=(\mathcal V, \mathcal E, T)$. In this graph, the node set $\mathcal V={v_1, \dots, v_N}$ represents $N$ individual reviews, and the edge set $\mathcal E$ captures relationships between these review nodes. Each review node $v_i \in \mathcal V$ is associated with a text $T_i$, which consists of multiple tokens $(T_{i1}, T_{i2}, \dots)$. Some of the nodes have known labels $y_i \in {0,1}$, where 0 indicates a normal review and 1 indicates a spam (fraudulent) review. Typically, only a small portion of nodes are labeled for training, while most remain unlabeled. The goal is to predict which of the unlabeled nodes are spam, using both the graph structure and the content of the reviews.

\subsection{The \fd Detector} 

Our detection model, \fd, consists of \textbf{\emph{four main components}}: review graph construction, language model (LM)-enhanced node embeddings, gated graph transformers, and MLP layers for classification. The overall architecture of \fd is shown in Figure \ref{fig:fraudsquad_architecture}.
Compared to prior work, \fd introduces \emph{two key innovations}. 
First, while state-of-the-art fraud detection methods, such as CARE-GNN \cite{Dou2020EnhancingGN}, PC-GNN \cite{liu2021pick}, DGA-GNN \cite{dgagnn}, and GTAN \cite{Xiang2023semi}, typically rely on feature engineering for node representation, \fd leverages advanced language models to \emph{directly process raw review texts}, enabling richer semantic understanding. 
Second, \fd adopts a more capable architecture by using \emph{gated graph transformers}. In contrast, prior approaches use message-passing and aggregation mechanisms (e.g., CARE-GNN, PC-GNN, and DGA-GNN) or graph attention networks tailored for temporal graphs (e.g., GTAN). In the following paragraphs, we describe the design and components of \fd in more detail.

\para{Review graph construction.}
The review graph is built to \emph{capture fraudster behavior}, specifically the act of users posting reviews on products, services, or other items. In typical review platforms like Amazon, we construct edges between review nodes based on three types of relationships, following established practices \cite{Dou2020EnhancingGN,liu2021pick}:
(1) connecting pairs of reviews written by the same \emph{user};
(2) connecting reviews of the same \emph{product} that share the same \emph{star rating};
(3) connecting reviews of the same \emph{product} posted within the same \emph{month}. 
This framework can be adapted to other scenarios as well. For example, in question-answering platforms where answers serve as reviews of the product mentioned in the question, the graph can be constructed using the following relationships:
(1) connecting QA pairs under the same \emph{question};
(2) connecting QA pairs where the \emph{questions} are asked by the same user within the same \emph{month};
(3) connecting QA pairs where the \emph{answers} are given by the same user within the same \emph{month}.

\para{LM-enhanced node embeddings.}
The LM-enhanced node embeddings are designed to \emph{capture the semantic features of review texts}. We construct the initial embeddings for each node in the review graph as follows.
First, each review text $T_i$ is passed through a pre-trained language model (LM) with frozen weights to obtain its text embedding $X_i$ in a latent space. 

Next, we derive a trainable risk embedding $Z_i$ based on the node labels to enable label propagation alongside feature propagation in the graph neural network. As per \citet{shimasked}, integrating label and feature (i.e., text embeddings) propagation within the same GNN framework improves performance. The risk embedding vocabulary includes three classes, \emph{normal}, \emph{fraud}, and \emph{unknown}, and the embedding dimensionality is set to match that of the text embedding. To avoid label leakage, in each training batch, the labels of the training nodes are all masked as \emph{unknown}, so that these nodes can only aggregate the risk embeddings from their neighbors. 

Finally, we combine the text embedding and the risk embedding using trainable weights $\beta_1, \beta_2, \beta_3$. The resulting initial node embedding $H_i$ is computed as:
\begin{equation}
	H_i = X_i + \text{PReLU}(X_i\beta_1 + Z_i\beta_2)\beta_3,
\end{equation}
where PReLU \cite{he2015delvingdeeprectifierssurpassing} is a non-linear activation function.

\para{Gated graph transformers.}
We input the initial node embeddings $H_i$ into gated graph transformer layers with multi-headed attention. 

The graph transformer architecture \cite{Dwivedi2020AGO} uses three fundamental embedding vectors for each node $v_i$: Query (Q), Key (K), and Value (V), computed as follows for each attention head $s = 1, \dots, S$: 
\begin{equation}
\begin{aligned}
	Q_{is} &= H_i W_{query}^s, \\
	V_{is} &= H_i W_{value}^s, \\
	K_{is} &= H_i W_{key}^s,\\
\end{aligned}
\end{equation}
where $W_{query}^s, W_{value}^s, W_{key}^s$ are learnable weights for the $s$-th attention head.
For a given node $v_i$, we denote its set of neighbors as $\mathcal{N}_i$. The attention coefficient between $v_i$ and a neighbor $v_j \in \mathcal{N}_i$ is calculated by measuring the similarity between their Q and K embeddings:
\begin{equation}
\alpha_{ij}^s = \frac{\exp(Q_{is}^\top K_{js})}{\sum_{v_j \in \mathcal{N}i} \exp(Q{is}^\top K_{js})}.
\end{equation}
These attention weights determine how much influence each neighboring node's Value embedding contributes to the updated representation of $v_i$:
\begin{equation}
	\tilde{H}_i^{s} = \sum_{v_j\in \mathcal N_i} \alpha_{ij}^s V_{js}.
\end{equation}
We then concatenate the outputs from all $S$ attention heads to obtain the aggregated representation:
\begin{equation}
	\tilde{H}_i = \text{Concat}({\tilde{H}}_i^1, \dots, {\tilde{H}}_i^S).
\end{equation}
Finally, we compute a shortcut projection from the linearly transformed input of the layer $O_i$ using a gate mechanism following \citet{Xiang2023semi}:  
\begin{equation}
\begin{aligned}	
	O_i &= H_i\beta_3,\\
	\text{gate}_i &= \text{Sigmoid}(\text{Concat} (O_i, \tilde{H}_i, O_i-\tilde{H}_i)\beta_4), \\
	\hat{H}_i &= \text{gate}_i O_i + (1-\text{gate}_i)\tilde{H}_i.\\
\end{aligned}
\end{equation}
The resulting $\hat{H}_i$ serves as the final output of one gated graph transformer layer.

\para{MLP for classification.} 
After passing through $L = 2$ gated graph transformer layers, each node embedding is fed into a multi-layer perceptron (MLP) to produce a probability score indicating whether the review is fraudulent (spam).

The entire \fd model is trained using the labeled nodes with a binary cross-entropy loss. Optimization is performed using the Adam optimizer \cite{kingma2014adam}.

%% file: 3experiment_sim.tex
\section{Experiments}

To validate the effectiveness of our proposed method, we conduct comprehensive experiments on all three LLM-generated and two additional human-written spam review datasets. The following section outlines our experimental setup, main results, ablation studies, and a discussion on feature engineering.

\subsection{Experiment Setup}

\para{Datasets.} 
We evaluate our model on the datasets listed in Table \ref{table:dataset-stats}. In addition to the three LLM-generated spam review datasets synthesized in Section \ref{sec:syn_data}, we also include two human-written spam review datasets.
The \dyelpres dataset \cite{Akoglu15} is a public collection of Yelp hotel reviews from Chicago, with labels provided by Yelp indicating whether each review is filtered (spam) or recommended (normal).
The \dqa dataset \cite{Liu2017Detecting} is sourced from a Chinese community question-answering platform in 2015. It contains both genuine QA pairs and manipulated content created by large-scale crowdsourcing campaigns, often used for brand promotion or misinformation. These manipulated entries are treated as spam for detection purposes.
All five datasets include ground-truth labels for evaluation.

\begin{table}[tb]
\centering
\caption{Statistics of datasets selected for experiments. The first three datasets are synthesized and contain LLM-generated spam reviews. The last two are publicly available datasets containing human-written spam reviews.}
\label{table:dataset-stats}
\begin{tabular}{@{}lrrrrr@{}}
\toprule
Dataset & Nodes & Edges & Spam nodes \\ \midrule
\texttt{Amazon-\llamathree}  & 89,186          & 4,139,448  & 2.8\% \\ 
\texttt{Amazon-\qwen}  & 89,192          & 4,140,166 & 2.8\% \\
\texttt{Amazon-\ds}  & 89,197          & 4,138,569 & 2.8\% \\ 
\cmidrule{1-4}
\dyelpres    & 5,854           & 141,123 & 13.3\% \\
\dqa  & 133,317     & 66,272,741  & 34.2\% \\  
\bottomrule
\end{tabular}
\end{table}

\begin{table*}[tb]
  \caption{Detection performance (\%) on both LLM-generated and human-written spam reviews. Pre: precision, Rec: recall.}
  \centering
  \setlength{\tabcolsep}{4.3pt} 
  \begin{tabular}{l|ccc|ccc|ccc|ccc|ccc}
  \toprule
  \multirow{3}{*}{Method}     &\multicolumn{9}{c|}{LLM-generated spam review datasets} &\multicolumn{6}{c}{Human-written spam review datasets}\\
  \cmidrule(lr){2-10}\cmidrule(lr){11-16}
  &\multicolumn{3}{c|}{\texttt{\dama-\qwen}}   & \multicolumn{3}{c|}{\texttt{\dama-\llamathree}}  & \multicolumn{3}{c|}{\texttt{\dama-{\ds}}} & \multicolumn{3}{c|}{\dyelpres} & \multicolumn{3}{c}{\dqa} \\\cmidrule(lr){2-4}\cmidrule(lr){5-7}\cmidrule(lr){8-10}\cmidrule(lr){11-13}\cmidrule(lr){14-16}
        & AUC & Prec & Rec & AUC & Prec & Rec & AUC & Prec & Rec & AUC & Prec & Rec & AUC & Prec & Rec\\\midrule
  MLP           & 79.37 & 35.26 & 37.80 & 69.72 & 8.64 & 9.29 & 92.50 & 45.93 & 49.16 & \underline{61.05} & 21.27 & 24.00 & 77.42 & 63.58 & 55.79\\ 
  RNN           & \underline{96.41} & \underline{48.14} & \underline{56.01} & 94.07 & 68.86 & 57.46 & 96.64 & \underline{75.52} & 58.13 & 55.78 & 12.60 & 2.29 & 58.53 & 32.74 & 5.31\\ 
  GAT \cite{graphattentionnetworks}           & 86.09 & 43.48 & 46.62 & 77.45 & 21.47 & 23.08 & 74.61 & 23.21 & 24.84 & 60.88 & 21.39 & 24.14 & 72.00 & 57.41 & 50.37\\ \cmidrule{1-16}
  CARE-GNN \cite{Dou2020EnhancingGN}      & 89.69 & 44.48 & 47.68 & 95.63 & 45.68 & 49.11 & 95.11 & 48.01 & 51.38 & 54.80 & 18.73 & 21.14 & 78.78 & 68.47 & 60.08\\  
  PC-GNN \cite{liu2021pick}        & 83.43 & 31.98 & 34.28 & 94.69 & 42.86 & 46.07 & 96.45 & 51.29 & 54.89 & 59.16 & \underline{22.28} & \underline{25.14} & 78.68 & 66.96 & 58.76\\  
  GTAN \cite{Xiang2023semi}          & 90.00 & 45.14 & 48.40 & \underline{96.91} & 62.87 & 67.59 & \underline{97.95} & 64.87 & \underline{69.42} & 60.25 & 21.27 & 24.00 & \underline{79.81} & \underline{69.63} & \underline{61.10}\\ 
  DGA-GNN \cite{dgagnn}       & 89.05 & 36.41 & 42.65 & 93.30 & \underline{78.23} & \underline{80.85} & 71.67 & 21.77 & 17.16 & 56.01 & 20.69 & 9.43 & 66.80 & 46.70 & 56.84\\ \cmidrule{1-16}
  \fd    & \textbf{99.98} & \textbf{92.36} & \textbf{99.02} & \textbf{99.94} & \textbf{90.99} & \textbf{97.81} & \textbf{99.93} & \textbf{89.45} & \textbf{95.73} & \textbf{70.32} & \textbf{33.67} & \textbf{38.00} & \textbf{99.43} & \textbf{99.91} & \textbf{87.67}\\ \bottomrule
  \end{tabular}
  \label{tab:performance_combined}
  \end{table*}

\para{Baselines.}
To demonstrate the effectiveness of our proposed hybrid spam review detection model \fd, we compare it against several strong baselines. Baselines (1)–(3) are general-purpose classification models, while baselines (4)–(7) are state-of-the-art GNN-based fraud detection methods:
(1) \emph{MLP}: A multi-layer perceptron with two hidden layers that takes numerical features as input.
(2) \emph{RNN}: A recurrent neural network that processes the raw review texts as input.
(3) \emph{GAT} \cite{graphattentionnetworks}: A graph attention network that performs node classification using node features and graph structure.
(4) \emph{CARE-GNN} \cite{Dou2020EnhancingGN}: A graph neural network designed to handle fraud camouflage by enhancing the aggregation process.
(5) \emph{PC-GNN} \cite{liu2021pick}: A GNN that addresses class imbalance issues commonly found in fraud detection tasks.
(6) \emph{GTAN} \cite{Xiang2023semi}: A gated temporal attention network developed for fraud detection, particularly in domains like credit card transactions.
(7) \emph{DGA-GNN} \cite{dgagnn}: A dynamic grouping aggregation GNN tailored for fraud detection scenarios.

For models that require numerical node features (e.g., MLP and the GNN-based baselines), we use the engineered features provided in the original works \cite{Akoglu15, Dou2020EnhancingGN}.
All baseline models, as well as \fd, are implemented in PyTorch.

\para{Training setups.}
We adopt \emph{a minimally supervised setting} in our experiments, reflecting the real-world challenge of obtaining labeled data for the spam review detection task \cite{Yu_Liu_Luo_2024}. The dataset splits of the training-validation-testing are set as 1\%-9\%-90\% for all datasets except \dqa. Given the larger scale of the \dqa dataset, we use an even more challenging split:  0.1\%-9.9\%-90\%. Besides, all methods, including \fd run on one GPU with 48 GB of memory, showing a resource-constrained training environment.

For \fd, we use BERT-base-uncased \cite{devlin2019bert} as the language model to generate text embeddings. The gated graph transformer layers are configured with a hidden dimension of 100 and 3 attention heads. Training is run for up to 50 epochs. The gated graph transformer architecture is implemented using the Deep Graph Library (DGL) \cite{wang2020deepgraphlibrarygraphcentric}.

\para{Evaluation metrics.}
We evaluate model performance using \emph{precision}, \emph{recall}, and \emph{AUC} (area under the ROC curve).
In specific, after the detection model assigns a probability score to each node indicating the likelihood of being spam, we identify the top-ranked nodes as predicted spam and compute precision and recall accordingly.
In practice, fraud detection systems often require manual verification to avoid removing genuine content. Therefore, it is important to control the number of flagged candidates. To reflect this, we set the top-ranked prediction ratios based on the approximate spam prevalence in each dataset: 3\% for all LLM-generated review spam datasets, 15\% for \dyelpres, and 30\% for \dqa. 

\subsection{Results on Detecting LLM-Generated Spam Reviews}
Table \ref{tab:performance_combined} presents the detection performance on the three synthetic datasets containing LLM-generated spam reviews. For each metric, the best-performing method is highlighted in \textbf{bold}, and the second-best is \underline{underlined}.
Across all metrics and datasets, \fd consistently achieves strong performance, with scores ranging from 89.45\% to 99.98\%, outperforming all baselines by a large extent. 
Though other methods, including RNN and DGA-GNN, could have relatively high AUC scores, their precision and recall scores are significantly low, which indicates that \textbf{\emph{the top suspicious review nodes they predict are not accurate}}. 
These results demonstrate that \fd can effectively and accurately detect LLM-generated spam reviews.
Under a challenging, minimally supervised setting and resource-constrained environment, \fd remains highly effective in countering LLM-driven spam attacks.  Notably, the language model used for node embeddings in \fd is smaller than the generative LLMs that produce the spam, leading to a \textbf{\emph{lower detection cost than the cost of generation}}. This cost advantage helps reduce the economic incentive for carrying out such review spamming attacks.

Moreover, the three LLMs used to generate spam reviews show noticeable differences in their ability to evade detection, particularly against graph-based detectors. Among them, \qwen and \ds exhibit stronger evasion capabilities compared to \llamathree. For instance, when detecting spam reviews generated by \qwen, GNN-based baselines achieve precision and recall scores barely above 50\%. In contrast, these same baselines typically achieve over 70\% precision and recall when identifying spam generated by the other two LLMs. Notably, \fd records its lowest precision and recall when detecting \ds-generated spam, further highlighting the evasiveness of \ds. These observations suggest that \qwen and \ds are more effective at generating hard-to-detect spam review texts.

\subsection{Results on Detecting Human-Written Spam Reviews}
In addition to detecting LLM-generated spam reviews, \fd also achieves the best performance on identifying human-written spam reviews, as shown in the last two datasets in Table \ref{tab:performance_combined}.
However, we observe that detecting spam in the \dyelpres dataset is significantly more challenging than in the \dqa dataset, a trend consistent across all methods. On \dyelpres, detection metrics rarely exceed 70\% and often fall below 20\%. In contrast, detection scores on \dqa are generally above 60\%, with \fd achieving over 90\%. One likely reason for this difference is that spam in the \dqa dataset exhibits clearer patterns—both in expressive features (e.g., user grades for askers and answerers) and in coherent spamming behaviors \cite{Liu2017Detecting}. These features, when used as input to simple models like MLPs, already enable reasonable performance (e.g., precision and recall above 50\%). Additionally, the ground-truth labels in \dyelpres are derived from more complex filtering mechanisms, making the spam signals more difficult to learn.

\begin{figure}[tb]
  \centering
  \subfigure[\dama-\texttt{\qwen}]{
    \includegraphics[height=2.7cm]{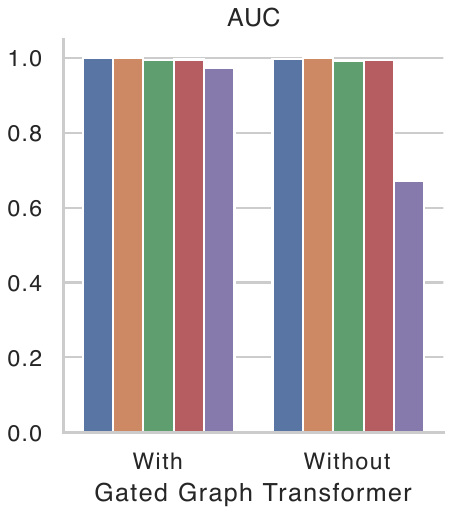}  
    \includegraphics[height=2.7cm]{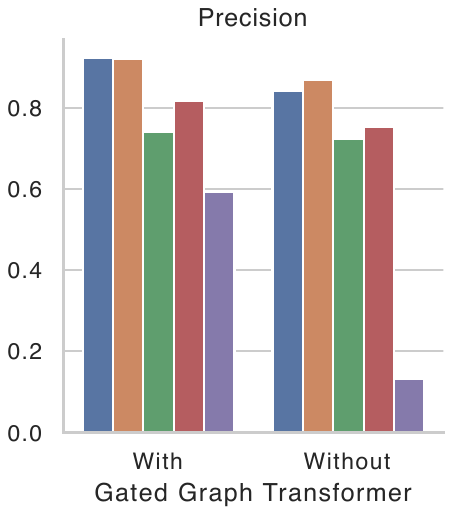}
	\includegraphics[height=2.7cm]{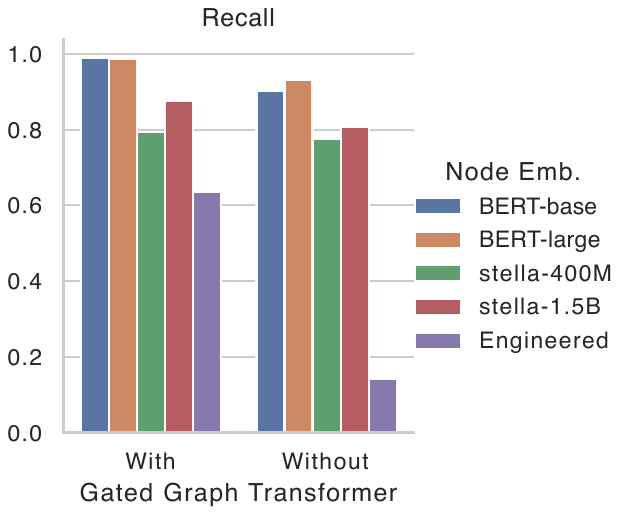} 
    \label{fig:qwen}
  }
  \subfigure[\dama-\texttt{\llamathree}]{
    \includegraphics[height=2.7cm]{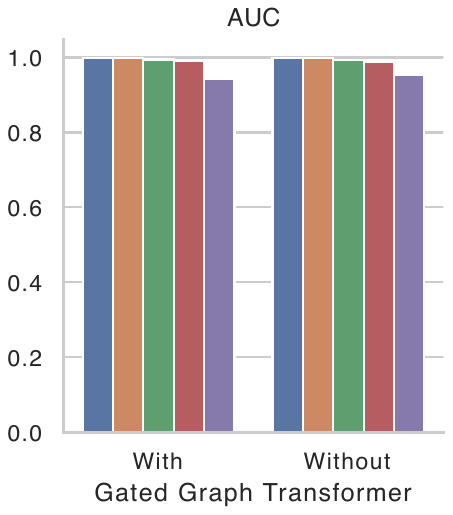}  
    \includegraphics[height=2.7cm]{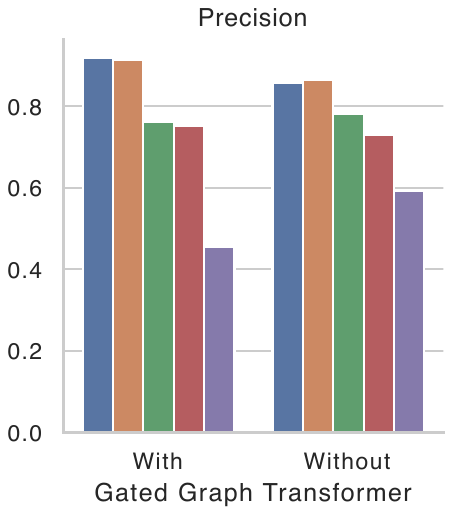}
	\includegraphics[height=2.7cm]{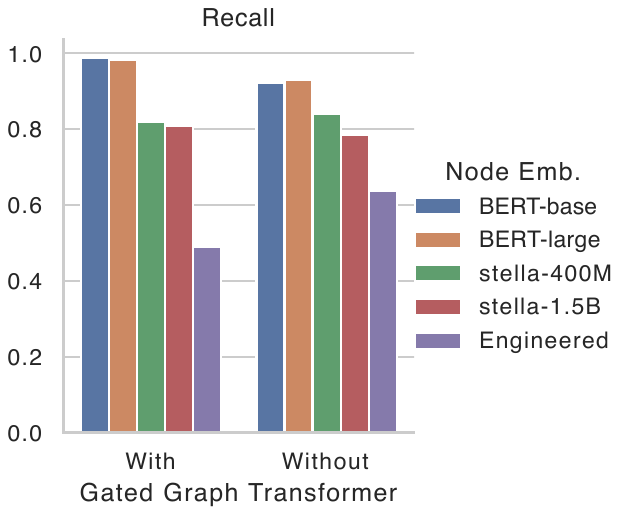}
    \label{fig:llama}
  }
  \subfigure[\dama-\texttt{\ds}]{
    \includegraphics[height=2.7cm]{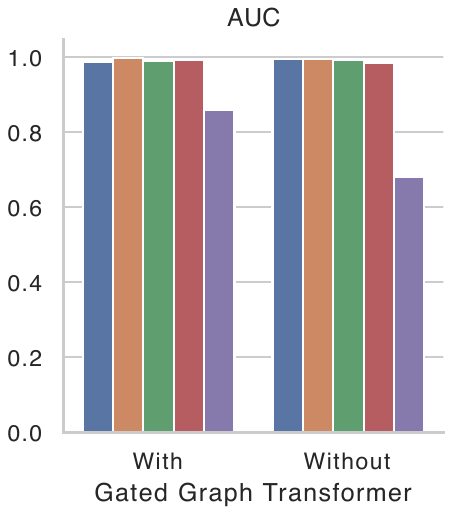}  
    \includegraphics[height=2.7cm]{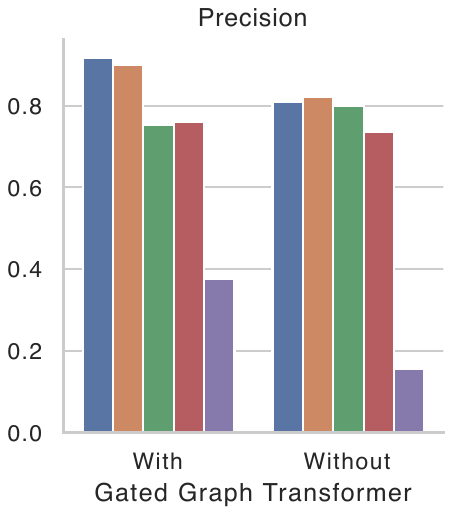}
	\includegraphics[height=2.7cm]{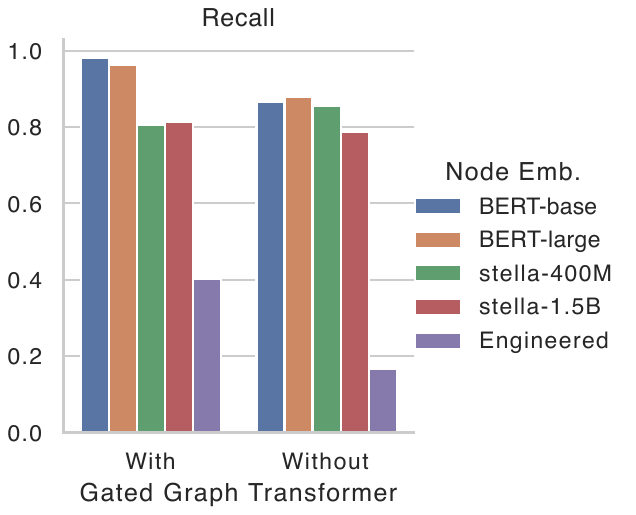}
    \label{fig:deepseek}
  }      
  \subfigure[\dyelpres]{
	\includegraphics[height=2.7cm]{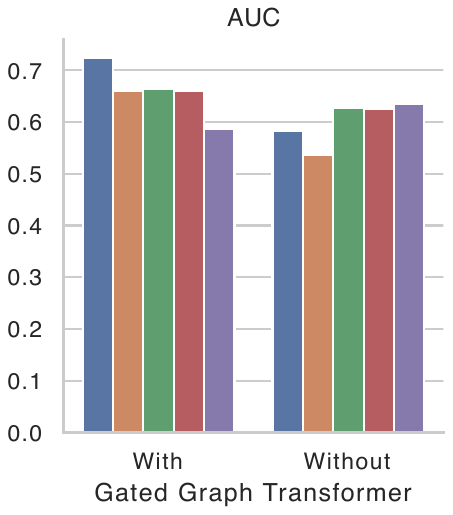}  
	\includegraphics[height=2.7cm]{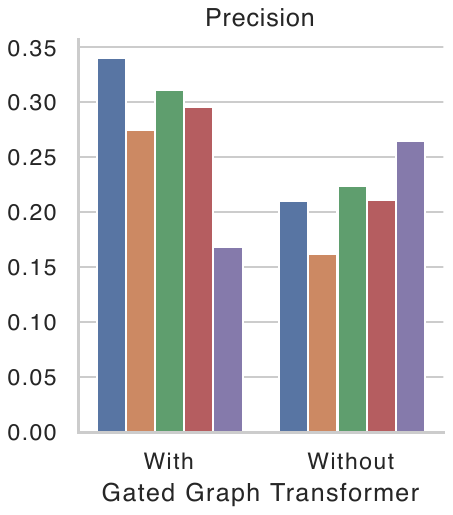}
	\includegraphics[height=2.7cm]{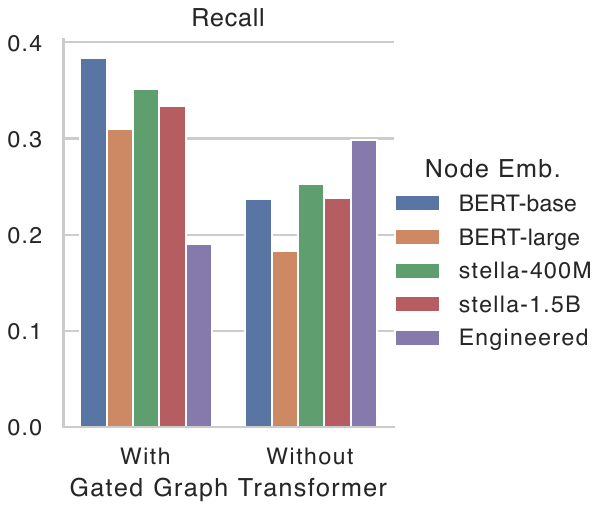}    
	\label{fig:yelp}
  }
  \subfigure[\dqa]{
	\includegraphics[height=2.7cm]{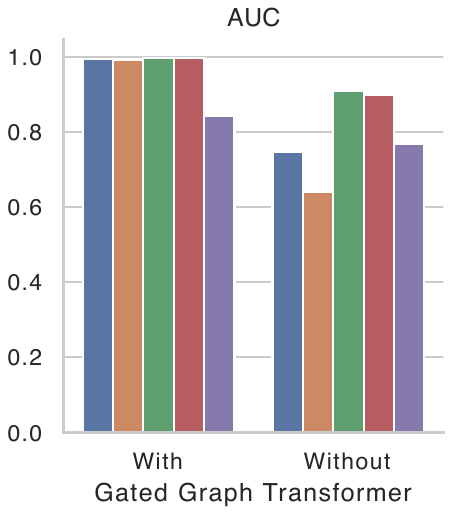}  
	\includegraphics[height=2.7cm]{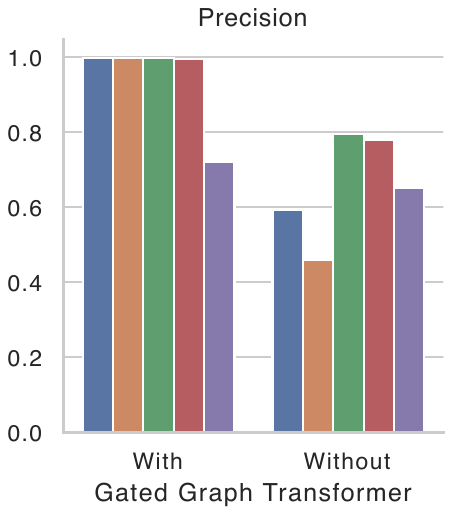}
	\includegraphics[height=2.7cm]{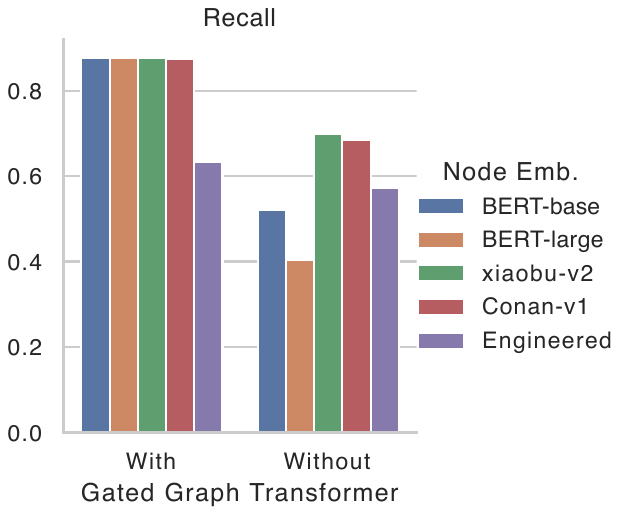}
    \label{fig:qa}
  }      
  \caption{Ablation study on node embeddings (different bars) and gated graph transformer (with vs. without sets of bars).}
  \label{fig:ablation_study}
\end{figure}

\begin{table*}[tb]
  \caption{Impact of engineered features (EF) on spam review detection performance. Scores for all metrics are shown in \%.}
  \centering
  \setlength{\tabcolsep}{4.3pt} 
  \begin{tabular}{l|ccc|ccc|ccc|ccc|ccc}
  \toprule
  \multirow{3}{*}{Setup}     &\multicolumn{9}{c|}{LLM-generated spam review datasets} &\multicolumn{6}{c}{Human-written spam review datasets}\\
  \cmidrule(lr){2-10}\cmidrule(lr){11-16}
  &\multicolumn{3}{c|}{\texttt{\dama-\qwen}}   & \multicolumn{3}{c|}{\texttt{\dama-\llamathree}}  & \multicolumn{3}{c|}{\texttt{\dama-{\ds}}} & \multicolumn{3}{c|}{\dyelpres} & \multicolumn{3}{c}{\dqa} \\\cmidrule(lr){2-4}\cmidrule(lr){5-7}\cmidrule(lr){8-10}\cmidrule(lr){11-13}\cmidrule(lr){14-16}
        & AUC & Prec & Rec & AUC & Prec & Rec & AUC & Prec & Rec & AUC & Prec & Rec & AUC & Prec & Rec \\\midrule
  With EF          & \textbf{99.99} & \textbf{92.82} & \textbf{99.51} & 99.51 & 85.67 & 92.10 & \textbf{99.90} & \textbf{91.99} & \textbf{98.44} & 68.85 & \textbf{32.41} & \textbf{36.57} & 99.37 & 99.83 & 87.60\\ 
  Without EF         & 99.97 & 92.69 & 99.38 & \textbf{99.98} & \textbf{91.82} & \textbf{98.71} & 99.72 & 90.32 & 96.67 & \textbf{70.11} & 31.65 & 35.71 & \textbf{99.39} & \textbf{99.94} & \textbf{87.70} \\\cmidrule{1-16}
  $\Delta$          & 0.02 & 0.13 & 0.13 & -0.47 & -6.15 & -6.61 & 0.18 & 1.67 & 1.77 & -1.26 & 0.76 & 0.86 & -0.02 & -0.11 & -0.10 \\ \bottomrule
  \end{tabular}
  \label{tab:performance_feature}
  \end{table*}

\subsection{Ablation Study on LM-Enhanced Node Embeddings}

We compare the effectiveness of LM-enhanced node embeddings with conventional feature-engineered node embeddings for spam review detection. We evaluate four language models for text embeddings: BERT-base and BERT-large \cite{devlin2019bert}, along with two recent embedding-focused LLMs—stella-400M and stella-1.5B \cite{zhang2025jasperstelladistillationsota}, both of which rank highly in the classification track of the Massive Text Embedding Benchmark (MTEB) \cite{muennighoff2022mteb}.
Considering \dqa is a Chinese dataset, we also include two Chinese embedding models: xiaobu-v2\footnote{\url{https://huggingface.co/lier007/xiaobu-embedding-v2}} and Conan-v1 \cite{li2024conanembeddinggeneraltextembedding}. All models except stella-1.5B are under 1B parameters in size. The detection results using these LMs are shown in Figure \ref{fig:ablation_study}.
Overall, LM-enhanced embeddings significantly outperform feature-engineered embeddings across all four datasets, particularly on the three LLM-generated spam review datasets. For the human-written spam datasets \dyelpres and \dqa, engineered features perform reasonably well in the absence of graph information (i.e., without gated graph transformers), but still fall short of the best-performing LM-based approaches.
Among the models tested, BERT remains a strong performer, striking a good balance between classification accuracy and model size. Notably, for the Chinese-language \dqa dataset, when no graph structure is used, the two Chinese embedding models outperform BERT, underscoring the importance of language-specific pretraining corpora.

\subsection{Ablation Study on Gated Graph Transformer}
While text embeddings produced by language models can be directly fed into a linear classifier with reasonably good accuracy, incorporating user behavior through the constructed review graph provides additional benefits. To evaluate this, we perform an ablation study by comparing detection performance with and without the use of gated graph transformers, as shown in Figure \ref{fig:ablation_study}. In the baseline setup, only text embeddings are used.
The results reveal a clear performance gap across all datasets. The improvement is especially pronounced for the human-written spam datasets, \dyelpres and \dqa, highlighting the importance of leveraging relational information captured in the review graph.

\subsection{Discussion: Are Engineered Features Still Necessary?}

Finally, we investigate whether the combination of LM-enhanced embeddings and graph neural networks effectively \emph{replaces the need for engineered features} that have historically played a key role in spam review detection. To explore this, we concatenate the engineered fraud features, derived from raw data by prior domain expertise, with the initial node embeddings before feeding them into the gated graph transformer layers. The performance differences are summarized in Table \ref{tab:performance_feature}.

Results show that adding engineered features has only a \emph{marginal} impact on detection performance, particularly for LLM-generated spam reviews. In many cases, the performance gap ($\Delta$) is negligible or even slightly negative. This suggests that the current architecture already captures the core patterns these features were designed to highlight.
That said, engineered features appear slightly more useful in detecting \emph{complex human-written spam}, such as the \dyelpres dataset. Therefore, we recommend that practitioners assess the relevance of such features based on the specific types of spam attacks they expect to encounter.
In general, we conclude that the integration of LM embeddings and graph-based modeling in \fd largely subsumes the value of engineered features. Nevertheless, \fd is flexible and can be easily \emph{augmented} to incorporate such features when needed.

%% file: 4related_work.tex
\section{Related Work}

Spam review detection has been widely studied, with research spanning from the design of meaningful features to the development of effective detection methods. Additionally, insights from misinformation detection offer valuable perspectives that can enhance spam detection approaches.

\para{Features of spam detection.} 
Numerous studies have explored the characteristics of fraudulent behavior, especially in the context of spam reviews \cite{Li2011Spam, Lim2010Detecting}. \citet{Rayana2015} provide a comprehensive list of engineered features commonly used for spam detection, forming the basis for many recent methods \cite{Dou2020EnhancingGN, liu2021pick}. There has been increasing interest in enriching feature sets by combining linguistic and behavioral signals, which often improves detection performance over using either alone. Integrating review graphs with metadata, capturing both textual and relational cues, has also been proven to be effective for more accurate and robust spam detection \cite{Rayana2015, Wang2017}. More recently, \citet{Xiang2023semi} improve detection effectiveness by jointly utilizing numerical and categorical attributes. \citet{dgagnn} further addresses the challenge of non-additive attributes by employing decision trees to encode non-additive node attributes into binarized vectors.

\para{Graph-based detection.}
Graph-based methods leverage the relationships among reviews, users, and products to enhance fraud detection. Early techniques like Graph Convolutional Networks (GCNs) \cite{gcn, hamilton2017inductive} have proven effective in capturing complex interactions within review graphs. Recent advances in Graph Neural Networks (GNNs) \cite{Dou2020EnhancingGN, Zhang2020, liu2021pick,RevisitingGraphBasedFraudDetection2023,Xiang2023semi,dgagnn} address key challenges such as class imbalance and heterophily. For example, spectral analysis has been integrated to enhance detection performance \cite{Zhang2020, RevisitingGraphBasedFraudDetection2023}. Methods like CARE-GNN and its enhanced variant RLC-GNN tackle issues of relation and feature camouflage, achieving notable gains in fraud detection accuracy \cite{RLCGNN2021}. Additionally, the Pick and Choose Graph Neural Network (PC-GNN) effectively mitigates class imbalance by selectively sampling nodes and edges to construct balanced subgraphs for training \cite{liu2021pick}. Furthermore, \citet{Xiang2023semi} propose a Gated Temporal Attention Network for fraud transaction detection, which is also applicable to spam review detection.

\para{Linguistic-based detection.}
These methods focus on analyzing the textual content of reviews to identify deceptive patterns. Common techniques include sentiment analysis, syntactic analysis, and lexical feature extraction. For instance, sentiment and psycholinguistic features have been incorporated to achieve higher detection accuracy, though these models often struggle with sophisticated fraudsters who mimic genuine review characteristics \cite{Ott2011, Li2014}. On the other hand, text embedding produced by language models are proposed to cover a range of tasks \cite{muennighoff2022mteb}, including retrieval, summarization, classification and more, which could be useful in detection.

\para{Misinformation detection.} Misinformation generated by LLMs has received increasing attention \cite{pan2023risk, preventing2024liu}. A notable example is the generation of targeted propaganda that closely imitates the style of legitimate news articles \cite{Zellers2019}.
DECOR \cite{DECOR23} is a recent method for fake news detection on the social graph. \citet{Wu2024Fake} also addresses the problem of fake news and proposes a  framework against style-based attacks from LLMs. \citet{feng2024does} is another work to detect bot accounts who utilize LLMs to evade detection in social media that post fake contents. While effective in the news domain, these approaches are specifically designed for that context and are not directly applicable to review-based scenarios.

%% file: 5conclusion.tex
\section{Conclusion}

In this work, we address the newly arisen challenge of detecting LLM-generated spam reviews by first synthesizing three realistic datasets that simulate how fraudsters might exploit LLMs to generate deceptive review content. Our analysis shows that these reviews are highly persuasive and closely resemble genuine human-written ones, motivating the urgency for robust detection methods.

To this end, we propose a novel hybrid spam detection approach, \fd, which integrates language model-enhanced node embeddings with gated graph transformers to jointly capture linguistic cues and user behavior patterns. Experimental results demonstrate that \fd is highly effective in detecting both LLM-generated and human-written spam reviews. These findings highlight the importance of combining semantic and structural signals to keep pace with increasingly sophisticated spam tactics.

\section*{Acknowledgement}
This work started from the project in the Deep Learning course in Tsinghua University. We sincerely thank Professor Xiaolin Hu and Professor Jun Zhu for their support.

%% file: 7appendix.tex
\appendix

\section{Detailed Prompts}

Table \ref{table:eval_prompt} presents the complete prompt used to evaluate five-star spam review texts, whether generated by \llms or written by humans. The prompt includes basic information about the target product, which is provided as input to \gpto.

\begin{table}[!htbp]
\centering
\caption{Prompts for evaluating spam review texts. Inputs of the prompt are highlighted using \textcolor{myblue2}{[]}.}
\label{table:eval_prompt}
\fontsize{8}{8}\selectfont
\begin{tabular}{p{1cm}p{6.5cm}}
\toprule
Role & Message\\ \midrule
\texttt{system} & You are a helpful assistant and know a lot about e-commerce on Amazon, especially about how the reviews influence potential customers. \\\midrule
\texttt{user} & Please first read a review about the product titled \textcolor{myblue2}{[product name]} in the category of \textcolor{myblue2}{[product category]}:\\
&\textcolor{myblue2}{[review to be evaluated]}\\
&Now, please evaluate the influence of the given review on a potential customer on Amazon in the following five aspects:\\
&- Will the user feel the review is positive? \\
&- Will the user feel the review contains useful details? \\
&- Will the user feel the review is convincing? \\
&- Will the user feel the review is written by a normal user? \\
&- Will the user be more willing to buy the product after reading the review? \\
&For each question, please first answer with a rating ranging from 1 (totally no) to 5 (totally yes) and then give a brief reason for the rating. \\
\bottomrule
\end{tabular}
\end{table}

\section{Qualitative Examples of LLM-Generated Spam Review Texts}
\label{appen:examples}

Tables \ref{table:llama3-review}, \ref{table:qwen-review}, and \ref{table:deepseek-review} provide qualitative examples of spam review texts generated by three distinct LLMs—\llamathree, \qwen, and \ds. These reviews target products from various categories within the \dama dataset.

\todo{Same product}
\begin{table}[!htbp]
\centering
\caption{Spam review texts generated by \llamathree.} 
\label{table:llama3-review}
\fontsize{7.5}{7.5}\selectfont
\begin{tabular}{p{1cm}p{7cm}}
\toprule
Category & Review text\\ \midrule
Musical Instruments & I'm not a professional musician, just a hobbyist who loves to play for fun. \patternd{I was blown away by} the quality of these strings - they brought out the best in my guitar and made it sound like a completely different instrument. The price is very reasonable too.\\\midrule
All Beauty & I'm \patternb{thrilled with} the results I've gotten from Scruples Quickseal Detangling Conditioner! As someone who color-treats their hair regularly, it's amazing to find a product that not only detangles but also helps maintain my color vibrancy. This stuff has become a regular in my beauty routine - can't recommend it enough! \\\midrule
Baby Products & \patternd{I was blown away by} this pet gate! As a busy pet parent, I need solutions that work quickly and easily. This magic gate exceeded my expectations - it's a breeze to set up, sturdy, and does exactly what I need it to do: keep my furry friend safe indoors. Plus, the design is sleek and modern, making it a great addition to my home.\\\midrule
Video Games & What a fantastic product! \patterna{I was skeptical at first, but} the 60-day pre-paid time card has given me the freedom to play without interruption. The in-game content is top-notch, and the community is so engaging. \patternc{A must-have for any} Star Wars fan like myself \\\midrule 
Software & What a treat! M\&M'S Adventure has become my go-to distraction when I need a break. The puzzles are cleverly designed, and the visuals are top-notch. I love how the game keeps me on my toes with new levels and characters to discover. It's also great that it's free, with no hidden costs or in-app purchases to worry about. \patternc{A must-have for any} puzzle game fan! \\
\bottomrule
\end{tabular}
\end{table}

\begin{table}[th]
\centering
\caption{Spam review texts generated by \qwen.} 
\label{table:qwen-review}
\fontsize{7.5}{7.5}\selectfont
\begin{tabular}{p{1cm}p{7cm}}
\toprule
Category & Review text \\ \midrule
All Beauty & Absolutely in love! I've never felt my skin so soft after using an exfoliating product. The 100\% PURE Cocoa Coffee Body Scrub is a game-changer. It feels like a spa treatment at home. The scent is divine, a warm and comforting aroma of cocoa that's not overpowering. The combination of sea salts and nourishing oils leaves my skin smooth as silk without any irritation. \patterne{Highly recommend} this for a luxurious pampering session! \\\midrule
Video Games & \patternb{Absolutely thrilled with my} purchase! The \$40 Xbox Gift Card was delivered instantly to my inbox, allowing me to download the latest games without any hassle. It's perfect for gamers who prefer digital copies and it’s also great when you want to treat yourself or a friend to some new adventures. \\\midrule
Baby Products & As a new mom, finding the right products for my baby has been essential. The KAKIBLIN Pillow is truly a lifesaver! Its soft fabric combined with organic pp cotton ensures that my little angel gets the best comfort possible. Even during long drives, he seems relaxed and happy. The fact that it's machine washable adds to its convenience. \patternc{A must-have for} parents! \\\midrule 
Software & \patternb{Absolutely thrilled with my} new Rii Wired Gaming Controller! It's super responsive and the dual-vibration/turbo feature really enhances my gaming experience on PC and Android. Love that it’s plug and play, so easy to set up. \\\midrule 
Health \& Personal Care & I've been using this shaver for a couple of weeks now, and I'm impressed by how well it performs every time. The cordless design adds so much flexibility in my routine. It's also great that the battery holds up really well between charges. \patternc{A must-have for} on-the-go grooming! \\
\bottomrule
\end{tabular}
\end{table}

\begin{table}[th]
\centering
\caption{Spam review texts generated by \ds.} 
\label{table:deepseek-review}
\fontsize{7.5}{7.5}\selectfont
\begin{tabular}{p{1cm}p{7cm}}
\toprule
Category & Review text \\ \midrule
All Beauty & This perfume exceeded my expectations! The blend of aromatic and woody notes creates a unique and captivating fragrance. It’s perfect for everyday use and leaves a subtle, sophisticated trail. The bottle is beautifully designed, and the spray application is smooth. A fantastic addition to my fragrance collection!  \\\midrule
Software & I’m thoroughly impressed with Wolf Quest! The high-quality 3D graphics and immersive gameplay make it a standout. I love the unlimited quests and the ability to grow a strong wolf pack. It’s a fantastic mix of adventure and strategy. This game has quickly become one of my favorites!  \\\midrule
Musical Instruments & This subwoofer is a game-changer for my home studio setup! The bass is deep, punchy, and incredibly clear. The build quality feels solid, and the adjustable crossover control allows me to fine-tune the sound to my liking. It’s perfect for both music production and casual listening. Highly recommend this for anyone looking to upgrade their audio experience.  \\\midrule
Baby Products & I absolutely love this gate! It’s lightweight and easy to install, which was a huge plus for me. The size is perfect for our doorway, and the mesh is sturdy enough to keep our curious puppy contained. It also looks great in our home, blending seamlessly with our decor. A must-have for any pet owner!  \\\midrule 
Appliances & This gap cover is exactly what I needed to keep my kitchen looking neat and tidy. The silicone material is flexible and non-slip, so it stays in place perfectly. I love that it’s easy to clean and can withstand high temperatures. It’s a small but incredibly useful product that makes a big difference. I’m so happy with my purchase!  \\\bottomrule
\end{tabular}
\end{table}

%% file: combined_ref.bib
@inproceedings{tianCFD,
  title={Crowd fraud detection in internet advertising},
  author={Tian, Tian and Zhu, Jun and Xia, Fen and Zhuang, Xin and Zhang, Tong},
  booktitle = {Proceedings of the 24th International Conference on World Wide Web},
  pages={1100--1110},
  year={2015},
}

@inproceedings{hooi2016Fraudar,
  title={FRAUDAR: Bounding Graph Fraud in the Face of Camouflage},
  author={Hooi, Bryan and Song, Hyun Ah and Beutel, Alex and Shah, Neil and Shin, Kijung and Faloutsos, Christos},
  booktitle = {Proceedings of the 22nd ACM SIGKDD International Conference on Knowledge Discovery and Data Mining},
  pages={895--904},
  year={2016},
}

@inproceedings{Xu2013CopyCatch,
  title={{CopyCatch}: Stopping Group Attacks by Spotting Lockstep Behavior In Social Networks},
  author={Beutel,Alex and Xu,Wanhong and Guruswami,Venkatesan and Palow,Christopher and Faloutsos,Christos},
  booktitle = {Proceedings of the 22nd International Conference on World Wide Web},
  pages={119-130},
  year={2013},
}

@inproceedings{Akoglu15,
author = {Shebuti Rayana and Leman Akoglu},
title = {Collective Opinion Spam Detection: Bridging Review Networks and metadata},
booktitle = {Proceeding of the 21st ACM SIGKDD international conference
on Knowledge discovery and data mining},
year = {2015},
}

@inproceedings{Zhang2020GCN,
author = {Zhang, Shijie and Yin, Hongzhi and Chen, Tong and Hung, Quoc Viet Nguyen and Huang, Zi and Cui, Lizhen},
title = {{GCN}-Based User Representation Learning for Unifying Robust Recommendation and Fraudster Detection},
year = {2020},
booktitle = {Proceedings of the 43rd International ACM SIGIR Conference on Research and Development in Information Retrieval},
pages = {689–698},
numpages = {10},
}

@article{Dou2020EnhancingGN,
title={Enhancing Graph Neural Network-based Fraud Detectors against Camouflaged Fraudsters},
author={Yingtong Dou and Zhiwei Liu and Li Sun and Yutong Deng and Hao Peng and Philip S. Yu},
journal={Proceedings of the 29th ACM International Conference on Information \& Knowledge Management},
year={2020}
}

@inproceedings{liu2021pick,
  title={Pick and Choose: A {GNN}-based Imbalanced Learning Approach for Fraud Detection},
  author={Liu, Yang and Ao, Xiang and Qin, Zidi and Chi, Jianfeng and Feng, Jinghua and Yang, Hao and He, Qing},
  booktitle={Proceedings of the Web Conference 2021},
  pages={3168--3177},
  year={2021}
}

@inproceedings{Jindal2008,
  title={Opinion spam and analysis},
  author={Jindal, Nitin and Liu, Bing},
  booktitle={Proceedings of the 2008 International Conference on Web Search and Data Mining},
  pages={219--230},
  year={2008},
}

@inproceedings{Ott2011,
  title={Finding deceptive opinion spam by any stretch of the imagination},
  author={Ott, Myle and Choi, Yejin and Cardie, Claire and Hancock, Jeffrey T},
  booktitle={Proceedings of the 49th Annual Meeting of the Association for Computational Linguistics: Human Language Technologies},
  pages={309--319},
  year={2011}
}

@inproceedings{Zellers2019,
  title={Defending against neural fake news},
  author={Zellers, Rowan and Holtzman, Ari and Bisk, Yonatan and Farhadi, Ali and Choi, Yejin},
  booktitle = {Proceedings of the 33rd International Conference on Neural Information Processing Systems}, 
  articleno = {812},
  volume={32},
  year={2019}
}

@inproceedings{Zhang2020,
  title={Fraud review detection using graph convolutional networks},
  author={Zhang, Yu and Tan, Pang-Ning and Ding, Ying},
  booktitle={Proceedings of the 29th ACM International Conference on Information and Knowledge Management},
  pages={2773--2781},
  year={2020},
  organization={ACM}
}

@article{Li2014,
  title={Towards a holistic approach to detect spam reviews in online review platforms},
  author={Li, Fei and Huang, Minlie and Yang, Yi and Zhu, Xiaoyan},
  journal={Proceedings of the 23rd International Conference on World Wide Web},
  pages={459--470},
  year={2014}
}

@article{RevisitingGraphBasedFraudDetection2023,
  title={Revisiting Graph-Based Fraud Detection in Sight of Heterophily and Spectrum},
  author={Xu, Fan and Wang, Nan and Wu, Hao and Wen, Xuezhi and Zhao, Xibin and Wan, Hai},
  journal={arXiv preprint arXiv:2312.06441}  ,
  year={2023}
}

@article{RLCGNN2021,
  title={{RLC-GNN}: An Improved Deep Architecture for Spatial-Based Graph Neural Network with Application to Fraud Detection},
  author={Zeng, Yufan and Tang, Jiashan},
  journal={Applied Sciences},
  volume={11},
  number={12},
  pages={5656},
  year={2021}
}

@inproceedings{Rayana2015,
  title={Collective opinion spam detection: Bridging review networks and metadata},
  author={Rayana, Shebuti and Akoglu, Leman},
  booktitle={Proceedings of the 21th ACM SIGKDD International Conference on Knowledge Discovery and Data Mining},
  pages={985--994},
  year={2015}
}

@inproceedings{Wang2017,
  title={Using a hybrid content-based and behavior-based featuring approach in fake review detection},
  author={Wang, Jian and Feng, Shuhua and Liu, Bing and Li, Yuming},
  booktitle={Proceedings of the 2017 International Conference on Information Systems},
  pages={849--861},
  year={2017}
}

@article{hou2024bridging,
  title={Bridging Language and Items for Retrieval and Recommendation},
  author={Hou, Yupeng and Li, Jiacheng and He, Zhankui and Yan, An and Chen, Xiusi and McAuley, Julian},
  journal={arXiv preprint arXiv:2403.03952},
  year={2024}
}

@inproceedings{graphattentionnetworks,
      title={Graph Attention Networks}, 
      author={Petar Veličković and Guillem Cucurull and Arantxa Casanova and Adriana Romero and Pietro Liò and Yoshua Bengio},
      year={2018},
      booktitle = {Proceedings of the Sixth International Conference on Learning Representations},
}

@misc{wang2020deepgraphlibrarygraphcentric,
      title={Deep Graph Library: A Graph-Centric, Highly-Performant Package for Graph Neural Networks}, 
      author={Minjie Wang and Da Zheng and Zihao Ye and Quan Gan and Mufei Li and Xiang Song and Jinjing Zhou and Chao Ma and Lingfan Yu and Yu Gai and Tianjun Xiao and Tong He and George Karypis and Jinyang Li and Zheng Zhang},
      year={2020},
      eprint={1909.01315},
      archivePrefix={arXiv},
}

@inproceedings{Liu2017Detecting,
author = {Liu, Yuli and Liu, Yiqun and Zhou, Ke and Zhang, Min and Ma, Shaoping},
title = {Detecting Collusive Spamming Activities in Community Question Answering},
year = {2017},
booktitle = {Proceedings of the 26th International Conference on World Wide Web},
pages = {1073–1082},
numpages = {10},
}

@article{qwen2,
      title={Qwen2 Technical Report}, 
      author={An Yang and Baosong Yang and Binyuan Hui and Bo Zheng and Bowen Yu and Chang Zhou and Chengpeng Li and Chengyuan Li and Dayiheng Liu and Fei Huang and Guanting Dong and Haoran Wei and Huan Lin and Jialong Tang and Jialin Wang and Jian Yang and Jianhong Tu and Jianwei Zhang and Jianxin Ma and Jin Xu and Jingren Zhou and Jinze Bai and Jinzheng He and Junyang Lin and Kai Dang and Keming Lu and Keqin Chen and Kexin Yang and Mei Li and Mingfeng Xue and Na Ni and Pei Zhang and Peng Wang and Ru Peng and Rui Men and Ruize Gao and Runji Lin and Shijie Wang and Shuai Bai and Sinan Tan and Tianhang Zhu and Tianhao Li and Tianyu Liu and Wenbin Ge and Xiaodong Deng and Xiaohuan Zhou and Xingzhang Ren and Xinyu Zhang and Xipin Wei and Xuancheng Ren and Yang Fan and Yang Yao and Yichang Zhang and Yu Wan and Yunfei Chu and Yuqiong Liu and Zeyu Cui and Zhenru Zhang and Zhihao Fan},
      journal={arXiv preprint arXiv:2407.10671},
      year={2024}
}

@inproceedings{Xiang2023semi,
author = {Xiang, Sheng and Zhu, Mingzhi and Cheng, Dawei and Li, Enxia and Zhao, Ruihui and Ouyang, Yi and Chen, Ling and Zheng, Yefeng},
title = {Semi-Supervised Credit Card Fraud Detection via Attribute-Driven Graph Representation},
year = {2023},
booktitle = {Proceedings of the Thirty-Seventh AAAI Conference on Artificial Intelligence},
numpages = {9},
articleno = {1633},
}

@inproceedings{dgagnn, 
title={{DGA-GNN}: Dynamic Grouping Aggregation {GNN} for Fraud Detection}, 
abstractNote={Fraud detection has increasingly become a prominent research field due to the dramatically increased incidents of fraud. The complex connections involving thousands, or even millions of nodes, present challenges for fraud detection tasks. Many researchers have developed various graph-based methods to detect fraud from these intricate graphs. However, those methods neglect two distinct characteristics of the fraud graph: the non-additivity of certain attributes and the distinguishability of grouped messages from neighbor nodes.
This paper introduces the Dynamic Grouping Aggregation Graph Neural Network (DGA-GNN) for fraud detection, which addresses these two characteristics by dynamically grouping attribute value ranges and neighbor nodes. In DGA-GNN, we initially propose the decision tree binning encoding to transform non-additive node attributes into bin vectors. This approach aligns well with the GNN’s aggregation operation and avoids nonsensical feature generation. Furthermore, we devise a feedback dynamic grouping strategy to classify graph nodes into two distinct groups and then employ a hierarchical aggregation. This method extracts more discriminative features for fraud detection tasks. Extensive experiments on five datasets suggest that our proposed method achieves a 3% ~ 16% improvement over existing SOTA methods. Code is available at https://github.com/AtwoodDuan/DGA-GNN.}, 
booktitle={Proceedings of the AAAI Conference on Artificial Intelligence}, 
author={Duan, Mingjiang and Zheng, Tongya and Gao, Yang and Wang, Gang and Feng, Zunlei and Wang, Xinyu}, 
year={2024}, 
pages={11820-11828} }

@inproceedings{DECOR23,
author = {Wu, Jiaying and Hooi, Bryan},
title = {{DECOR}: Degree-Corrected Social Graph Refinement for Fake News Detection},
year = {2023},
booktitle = {Proceedings of the 29th ACM SIGKDD Conference on Knowledge Discovery and Data Mining},
pages = {2582–2593},
numpages = {12},
}

@inproceedings{Li2011Spam,
author = {Li, Fangtao and Huang, Minlie and Yang, Yi and Zhu, Xiaoyan},
title = {Learning to identify review spam},
year = {2011},
isbn = {9781577355151},
publisher = {AAAI Press},
booktitle = {Proceedings of the Twenty-Second International Joint Conference on Artificial Intelligence - Volume Volume Three},
pages = {2488–2493},
numpages = {6},
location = {Barcelona, Catalonia, Spain},
series = {IJCAI'11}
}

@inproceedings{Lim2010Detecting,
author = {Lim, Ee-Peng and Nguyen, Viet-An and Jindal, Nitin and Liu, Bing and Lauw, Hady Wirawan},
title = {Detecting product review spammers using rating behaviors},
year = {2010},
isbn = {9781450300995},
publisher = {Association for Computing Machinery},
address = {New York, NY, USA},
url = {https://doi.org/10.1145/1871437.1871557},
doi = {10.1145/1871437.1871557},
booktitle = {Proceedings of the 19th ACM International Conference on Information and Knowledge Management},
pages = {939–948},
numpages = {10},
location = {Toronto, ON, Canada},
series = {CIKM '10}
}

@inproceedings{caoSynchroTrap,
  title={Uncovering Large Groups of Active Malicious Accounts in Online Social Networks},
  author={Cao, Qiang and Yang, Xiaowei and Yu, Jieqi and Palow, Christopher},
  booktitle = {Proceedings of the 2014 ACM SIGSAC Conference on Computer and Communications Security},
  pages={477--488},
  year={2014},
  organization={ACM}
}

@inproceedings{gcn,
  title={Semi-Supervised Classification with Graph Convolutional Networks},
  author={Kipf, Thomas N. and Welling, Max},
  booktitle={Proceedings of the International Conference on Learning Representations},
  year={2017},
}

@inproceedings{hamilton2017inductive,
  title={Inductive Representation Learning on Large Graphs},
  author={Hamilton, Will and Ying, Zhitao and Leskovec, Jure},
  booktitle={Advances in Neural Information Processing Systems},
  pages={1024--1034},
  year={2017}
}

@article{Yu_Liu_Luo_2024, 
title={Barely Supervised Learning for Graph-Based Fraud Detection}, 
journal={Proceedings of the AAAI Conference on Artificial Intelligence}, 
author={Yu, Hang and Liu, Zhengyang and Luo, Xiangfeng}, 
year={2024}, 
month={Mar.}, 
pages={16548-16557} 
}

@inproceedings{devlin2019bert,
  title={{BERT}: Pre-training of Deep Bidirectional Transformers for Language Understanding},
  author={Devlin, Jacob and Chang, Ming-Wei and Lee, Kenton and Toutanova, Kristina},
  booktitle={Proceedings of the 2019 Conference of the North American Chapter of the Association for Computational Linguistics: Human Language Technologies}, 
  year={2019},
}

@article{openai2023gpt4,
  title={GPT-4 Technical Report},
  author={OpenAI},
  journal={OpenAI Report},
  year={2023},
  url={https://cdn.openai.com/papers/gpt-4.pdf}
}

@article{Dwivedi2020AGO,
  title={A generalization of transformer networks to graphs},
  author={Dwivedi, Vijay Prakash and Bresson, Xavier},
  journal={arXiv preprint arXiv:2012.09699},
  year={2020}
}

@inproceedings{shimasked,
  title={Masked Label Prediction: Unified Message Passing Model for Semi-Supervised Classification},
  author={Shi, Yunsheng and Huang, Zhengjie and Feng, Shikun and Zhong, Hui and Wang, Wenjing and Sun, Yu},
    booktitle={Proceedings of the 30th International Joint Conference on Artificial Intelligence}, 
  year={2021},
}

@article{muennighoff2022mteb,
  url = {https://arxiv.org/abs/2210.07316},
  author = {Muennighoff, Niklas and Tazi, Nouamane and Magne, Lo{\"\i}c and Reimers, Nils},
  title = {{MTEB}: Massive Text Embedding Benchmark},
  publisher = {arXiv},
  journal={arXiv preprint arXiv:2210.07316},  
  year = {2022}
}

@inproceedings{Andresini2022ReviewSD,
  title={Review Spam Detection using Multi-View Deep Learning Combining Content and Behavioral Features},
  author={Giuseppina Andresini and Andrea Iovine and Roberto Gasbarro and Marco Lomolino and Marco Degemmis and Annalisa Appice},
  booktitle={The 1st Italian Conference on Big Data and Data Science (itaDATA)},
  year={2022},
}

@article{Duma2023ADH,
  title={A Deep Hybrid Model for fake review detection by jointly leveraging review text, overall ratings, and aspect ratings},
  author={Ramadhani Ally Duma and Zhendong Niu and Ally S. Nyamawe and Jude Tchaye-Kondi and Abdulganiyu Abdu Yusuf},
  journal={Soft Computing},
  year={2023},
  volume={27},
  pages={6281-6296},
}

@inproceedings{Wu2024Fake,
author = {Wu, Jiaying and Guo, Jiafeng and Hooi, Bryan},
title = {Fake News in Sheep's Clothing: Robust Fake News Detection Against LLM-Empowered Style Attacks},
year = {2024},
booktitle = {Proceedings of the 30th ACM SIGKDD Conference on Knowledge Discovery and Data Mining},
pages = {3367–3378},
numpages = {12},
}

@misc{zhang2025jasperstelladistillationsota,
      title={{Jasper and Stella}: distillation of {SOTA} embedding models}, 
      author={Dun Zhang and Jiacheng Li and Ziyang Zeng and Fulong Wang},
      year={2025},
      eprint={2412.19048},
      archivePrefix={arXiv}}

@misc{li2024conanembeddinggeneraltextembedding,
  title={Conan-embedding: General Text Embedding with More and Better Negative Samples}, 
  author={Shiyu Li and Yang Tang and Shizhe Chen and Xi Chen},
  year={2024},
  eprint={2408.15710},
  archivePrefix={arXiv},
}

@misc{deepseekai2025,
      title={{DeepSeek-R1}: Incentivizing Reasoning Capability in LLMs via Reinforcement Learning}, 
      author={DeepSeek-AI and Daya Guo and Dejian Yang and Haowei Zhang and Junxiao Song and Ruoyu Zhang and Runxin Xu and Qihao Zhu and Shirong Ma and Peiyi Wang and Xiao Bi and Xiaokang Zhang and Xingkai Yu and Yu Wu and Z. F. Wu and Zhibin Gou and Zhihong Shao and Zhuoshu Li and Ziyi Gao and Aixin Liu and Bing Xue and Bingxuan Wang and Bochao Wu and Bei Feng and Chengda Lu and Chenggang Zhao and Chengqi Deng and Chenyu Zhang and Chong Ruan and Damai Dai and Deli Chen and Dongjie Ji and Erhang Li and Fangyun Lin and Fucong Dai and Fuli Luo and Guangbo Hao and Guanting Chen and Guowei Li and H. Zhang and Han Bao and Hanwei Xu and Haocheng Wang and Honghui Ding and Huajian Xin and Huazuo Gao and Hui Qu and Hui Li and Jianzhong Guo and Jiashi Li and Jiawei Wang and Jingchang Chen and Jingyang Yuan and Junjie Qiu and Junlong Li and J. L. Cai and Jiaqi Ni and Jian Liang and Jin Chen and Kai Dong and Kai Hu and Kaige Gao and Kang Guan and Kexin Huang and Kuai Yu and Lean Wang and Lecong Zhang and Liang Zhao and Litong Wang and Liyue Zhang and Lei Xu and Leyi Xia and Mingchuan Zhang and Minghua Zhang and Minghui Tang and Meng Li and Miaojun Wang and Mingming Li and Ning Tian and Panpan Huang and Peng Zhang and Qiancheng Wang and Qinyu Chen and Qiushi Du and Ruiqi Ge and Ruisong Zhang and Ruizhe Pan and Runji Wang and R. J. Chen and R. L. Jin and Ruyi Chen and Shanghao Lu and Shangyan Zhou and Shanhuang Chen and Shengfeng Ye and Shiyu Wang and Shuiping Yu and Shunfeng Zhou and Shuting Pan and S. S. Li and Shuang Zhou and Shaoqing Wu and Shengfeng Ye and Tao Yun and Tian Pei and Tianyu Sun and T. Wang and Wangding Zeng and Wanjia Zhao and Wen Liu and Wenfeng Liang and Wenjun Gao and Wenqin Yu and Wentao Zhang and W. L. Xiao and Wei An and Xiaodong Liu and Xiaohan Wang and Xiaokang Chen and Xiaotao Nie and Xin Cheng and Xin Liu and Xin Xie and Xingchao Liu and Xinyu Yang and Xinyuan Li and Xuecheng Su and Xuheng Lin and X. Q. Li and Xiangyue Jin and Xiaojin Shen and Xiaosha Chen and Xiaowen Sun and Xiaoxiang Wang and Xinnan Song and Xinyi Zhou and Xianzu Wang and Xinxia Shan and Y. K. Li and Y. Q. Wang and Y. X. Wei and Yang Zhang and Yanhong Xu and Yao Li and Yao Zhao and Yaofeng Sun and Yaohui Wang and Yi Yu and Yichao Zhang and Yifan Shi and Yiliang Xiong and Ying He and Yishi Piao and Yisong Wang and Yixuan Tan and Yiyang Ma and Yiyuan Liu and Yongqiang Guo and Yuan Ou and Yuduan Wang and Yue Gong and Yuheng Zou and Yujia He and Yunfan Xiong and Yuxiang Luo and Yuxiang You and Yuxuan Liu and Yuyang Zhou and Y. X. Zhu and Yanhong Xu and Yanping Huang and Yaohui Li and Yi Zheng and Yuchen Zhu and Yunxian Ma and Ying Tang and Yukun Zha and Yuting Yan and Z. Z. Ren and Zehui Ren and Zhangli Sha and Zhe Fu and Zhean Xu and Zhenda Xie and Zhengyan Zhang and Zhewen Hao and Zhicheng Ma and Zhigang Yan and Zhiyu Wu and Zihui Gu and Zijia Zhu and Zijun Liu and Zilin Li and Ziwei Xie and Ziyang Song and Zizheng Pan and Zhen Huang and Zhipeng Xu and Zhongyu Zhang and Zhen Zhang},
      year={2025},
      eprint={2501.12948},
      archivePrefix={arXiv},
      primaryClass={cs.CL},
      url={https://arxiv.org/abs/2501.12948}, 
}

@article{zhao2023survey,
  title={A survey of large language models},
  author={Zhao, Wayne Xin and Zhou, Kun and Li, Junyi and Tang, Tianyi and Wang, Xiaolei and Hou, Yupeng and Min, Yingqian and Zhang, Beichen and Zhang, Junjie and Dong, Zican and others},
  journal={arXiv preprint arXiv:2303.18223},
  volume={1},
  number={2},
  year={2023}
}

@article{touvron2023llama,
  title={Llama: Open and efficient foundation language models},
  author={Touvron, Hugo and Lavril, Thibaut and Izacard, Gautier and Martinet, Xavier and Lachaux, Marie-Anne and Lacroix, Timoth{\'e}e and Rozi{\`e}re, Baptiste and Goyal, Naman and Hambro, Eric and Azhar, Faisal and others},
  journal={arXiv preprint arXiv:2302.13971},
  year={2023}
}

@misc{touvron2023llama2,
      title={Llama 2: Open Foundation and Fine-Tuned Chat Models}, 
      author={Hugo Touvron and Louis Martin and Kevin Stone and Peter Albert and Amjad Almahairi and Yasmine Babaei and Nikolay Bashlykov and Soumya Batra and Prajjwal Bhargava and Shruti Bhosale and Dan Bikel and Lukas Blecher and Cristian Canton Ferrer and Moya Chen and Guillem Cucurull and David Esiobu and Jude Fernandes and Jeremy Fu and Wenyin Fu and Brian Fuller and Cynthia Gao and Vedanuj Goswami and Naman Goyal and Anthony Hartshorn and Saghar Hosseini and Rui Hou and Hakan Inan and Marcin Kardas and Viktor Kerkez and Madian Khabsa and Isabel Kloumann and Artem Korenev and Punit Singh Koura and Marie-Anne Lachaux and Thibaut Lavril and Jenya Lee and Diana Liskovich and Yinghai Lu and Yuning Mao and Xavier Martinet and Todor Mihaylov and Pushkar Mishra and Igor Molybog and Yixin Nie and Andrew Poulton and Jeremy Reizenstein and Rashi Rungta and Kalyan Saladi and Alan Schelten and Ruan Silva and Eric Michael Smith and Ranjan Subramanian and Xiaoqing Ellen Tan and Binh Tang and Ross Taylor and Adina Williams and Jian Xiang Kuan and Puxin Xu and Zheng Yan and Iliyan Zarov and Yuchen Zhang and Angela Fan and Melanie Kambadur and Sharan Narang and Aurelien Rodriguez and Robert Stojnic and Sergey Edunov and Thomas Scialom},
      year={2023},
      eprint={2307.09288},
      archivePrefix={arXiv},
      primaryClass={cs.CL},
      url={https://arxiv.org/abs/2307.09288}, 
}

@inproceedings{preventing2024liu,
author = {Liu, Aiwei and Sheng, Qiang and Hu, Xuming},
title = {Preventing and Detecting Misinformation Generated by Large Language Models},
year = {2024},
booktitle = {Proceedings of the 47th International ACM SIGIR Conference on Research and Development in Information Retrieval},
pages = {3001–3004},
numpages = {4},
}

@article{pan2023risk,
  title={On the risk of misinformation pollution with large language models},
  author={Pan, Yikang and Pan, Liangming and Chen, Wenhu and Nakov, Preslav and Kan, Min-Yen and Wang, William Yang},
  journal={arXiv preprint arXiv:2305.13661},
  year={2023}
}

@inproceedings{feng2024does,
  title={What does the bot say? {Opportunities}m and risks of large language models in social media bot detection},
  author={Feng, Shangbin and Wan, Herun and Wang, Ningnan and Tan, Zhaoxuan and Luo, Minnan and Tsvetkov, Yulia},
  booktitle = "Proceedings of the 62nd Annual Meeting of the Association for Computational Linguistics",
  year={2024},
  pages = "3580--3601",
}

@misc{grattafiori2024llama3herdmodels,
      title={The Llama 3 Herd of Models}, 
      author={Aaron Grattafiori and Abhimanyu Dubey and Abhinav Jauhri and Abhinav Pandey and Abhishek Kadian and Ahmad Al-Dahle and Aiesha Letman and Akhil Mathur and Alan Schelten and Alex Vaughan and Amy Yang and Angela Fan and Anirudh Goyal and Anthony Hartshorn and Aobo Yang and Archi Mitra and Archie Sravankumar and Artem Korenev and Arthur Hinsvark and Arun Rao and Aston Zhang and Aurelien Rodriguez and Austen Gregerson and Ava Spataru and Baptiste Roziere and Bethany Biron and Binh Tang and Bobbie Chern and Charlotte Caucheteux and Chaya Nayak and Chloe Bi and Chris Marra and Chris McConnell and Christian Keller and Christophe Touret and Chunyang Wu and Corinne Wong and Cristian Canton Ferrer and Cyrus Nikolaidis and Damien Allonsius and Daniel Song and Danielle Pintz and Danny Livshits and Danny Wyatt and David Esiobu and Dhruv Choudhary and Dhruv Mahajan and Diego Garcia-Olano and Diego Perino and Dieuwke Hupkes and Egor Lakomkin and Ehab AlBadawy and Elina Lobanova and Emily Dinan and Eric Michael Smith and Filip Radenovic and Francisco Guzmán and Frank Zhang and Gabriel Synnaeve and Gabrielle Lee and Georgia Lewis Anderson and Govind Thattai and Graeme Nail and Gregoire Mialon and Guan Pang and Guillem Cucurell and Hailey Nguyen and Hannah Korevaar and Hu Xu and Hugo Touvron and Iliyan Zarov and Imanol Arrieta Ibarra and Isabel Kloumann and Ishan Misra and Ivan Evtimov and Jack Zhang and Jade Copet and Jaewon Lee and Jan Geffert and Jana Vranes and Jason Park and Jay Mahadeokar and Jeet Shah and Jelmer van der Linde and Jennifer Billock and Jenny Hong and Jenya Lee and Jeremy Fu and Jianfeng Chi and Jianyu Huang and Jiawen Liu and Jie Wang and Jiecao Yu and Joanna Bitton and Joe Spisak and Jongsoo Park and Joseph Rocca and Joshua Johnstun and Joshua Saxe and Junteng Jia and Kalyan Vasuden Alwala and Karthik Prasad and Kartikeya Upasani and Kate Plawiak and Ke Li and Kenneth Heafield and Kevin Stone and Khalid El-Arini and Krithika Iyer and Kshitiz Malik and Kuenley Chiu and Kunal Bhalla and Kushal Lakhotia and Lauren Rantala-Yeary and Laurens van der Maaten and Lawrence Chen and Liang Tan and Liz Jenkins and Louis Martin and Lovish Madaan and Lubo Malo and Lukas Blecher and Lukas Landzaat and Luke de Oliveira and Madeline Muzzi and Mahesh Pasupuleti and Mannat Singh and Manohar Paluri and Marcin Kardas and Maria Tsimpoukelli and Mathew Oldham and Mathieu Rita and Maya Pavlova and Melanie Kambadur and Mike Lewis and Min Si and Mitesh Kumar Singh and Mona Hassan and Naman Goyal and Narjes Torabi and Nikolay Bashlykov and Nikolay Bogoychev and Niladri Chatterji and Ning Zhang and Olivier Duchenne and Onur Çelebi and Patrick Alrassy and Pengchuan Zhang and Pengwei Li and Petar Vasic and Peter Weng and Prajjwal Bhargava and Pratik Dubal and Praveen Krishnan and Punit Singh Koura and Puxin Xu and Qing He and Qingxiao Dong and Ragavan Srinivasan and Raj Ganapathy and Ramon Calderer and Ricardo Silveira Cabral and Robert Stojnic and Roberta Raileanu and Rohan Maheswari and Rohit Girdhar and Rohit Patel and Romain Sauvestre and Ronnie Polidoro and Roshan Sumbaly and Ross Taylor and Ruan Silva and Rui Hou and Rui Wang and Saghar Hosseini and Sahana Chennabasappa and Sanjay Singh and Sean Bell and Seohyun Sonia Kim and Sergey Edunov and Shaoliang Nie and Sharan Narang and Sharath Raparthy and Sheng Shen and Shengye Wan and Shruti Bhosale and Shun Zhang and Simon Vandenhende and Soumya Batra and Spencer Whitman and Sten Sootla and Stephane Collot and Suchin Gururangan and Sydney Borodinsky and Tamar Herman and Tara Fowler and Tarek Sheasha and Thomas Georgiou and Thomas Scialom and Tobias Speckbacher and Todor Mihaylov and Tong Xiao and Ujjwal Karn and Vedanuj Goswami and Vibhor Gupta and Vignesh Ramanathan and Viktor Kerkez and Vincent Gonguet and Virginie Do and Vish Vogeti and Vítor Albiero and Vladan Petrovic and Weiwei Chu and Wenhan Xiong and Wenyin Fu and Whitney Meers and Xavier Martinet and Xiaodong Wang and Xiaofang Wang and Xiaoqing Ellen Tan and Xide Xia and Xinfeng Xie and Xuchao Jia and Xuewei Wang and Yaelle Goldschlag and Yashesh Gaur and Yasmine Babaei and Yi Wen and Yiwen Song and Yuchen Zhang and Yue Li and Yuning Mao and Zacharie Delpierre Coudert and Zheng Yan and Zhengxing Chen and Zoe Papakipos and Aaditya Singh and Aayushi Srivastava and Abha Jain and Adam Kelsey and Adam Shajnfeld and Adithya Gangidi and Adolfo Victoria and Ahuva Goldstand and Ajay Menon and Ajay Sharma and Alex Boesenberg and Alexei Baevski and Allie Feinstein and Amanda Kallet and Amit Sangani and Amos Teo and Anam Yunus and Andrei Lupu and Andres Alvarado and Andrew Caples and Andrew Gu and Andrew Ho and Andrew Poulton and Andrew Ryan and Ankit Ramchandani and Annie Dong and Annie Franco and Anuj Goyal and Aparajita Saraf and Arkabandhu Chowdhury and Ashley Gabriel and Ashwin Bharambe and Assaf Eisenman and Azadeh Yazdan and Beau James and Ben Maurer and Benjamin Leonhardi and Bernie Huang and Beth Loyd and Beto De Paola and Bhargavi Paranjape and Bing Liu and Bo Wu and Boyu Ni and Braden Hancock and Bram Wasti and Brandon Spence and Brani Stojkovic and Brian Gamido and Britt Montalvo and Carl Parker and Carly Burton and Catalina Mejia and Ce Liu and Changhan Wang and Changkyu Kim and Chao Zhou and Chester Hu and Ching-Hsiang Chu and Chris Cai and Chris Tindal and Christoph Feichtenhofer and Cynthia Gao and Damon Civin and Dana Beaty and Daniel Kreymer and Daniel Li and David Adkins and David Xu and Davide Testuggine and Delia David and Devi Parikh and Diana Liskovich and Didem Foss and Dingkang Wang and Duc Le and Dustin Holland and Edward Dowling and Eissa Jamil and Elaine Montgomery and Eleonora Presani and Emily Hahn and Emily Wood and Eric-Tuan Le and Erik Brinkman and Esteban Arcaute and Evan Dunbar and Evan Smothers and Fei Sun and Felix Kreuk and Feng Tian and Filippos Kokkinos and Firat Ozgenel and Francesco Caggioni and Frank Kanayet and Frank Seide and Gabriela Medina Florez and Gabriella Schwarz and Gada Badeer and Georgia Swee and Gil Halpern and Grant Herman and Grigory Sizov and Guangyi and Zhang and Guna Lakshminarayanan and Hakan Inan and Hamid Shojanazeri and Han Zou and Hannah Wang and Hanwen Zha and Haroun Habeeb and Harrison Rudolph and Helen Suk and Henry Aspegren and Hunter Goldman and Hongyuan Zhan and Ibrahim Damlaj and Igor Molybog and Igor Tufanov and Ilias Leontiadis and Irina-Elena Veliche and Itai Gat and Jake Weissman and James Geboski and James Kohli and Janice Lam and Japhet Asher and Jean-Baptiste Gaya and Jeff Marcus and Jeff Tang and Jennifer Chan and Jenny Zhen and Jeremy Reizenstein and Jeremy Teboul and Jessica Zhong and Jian Jin and Jingyi Yang and Joe Cummings and Jon Carvill and Jon Shepard and Jonathan McPhie and Jonathan Torres and Josh Ginsburg and Junjie Wang and Kai Wu and Kam Hou U and Karan Saxena and Kartikay Khandelwal and Katayoun Zand and Kathy Matosich and Kaushik Veeraraghavan and Kelly Michelena and Keqian Li and Kiran Jagadeesh and Kun Huang and Kunal Chawla and Kyle Huang and Lailin Chen and Lakshya Garg and Lavender A and Leandro Silva and Lee Bell and Lei Zhang and Liangpeng Guo and Licheng Yu and Liron Moshkovich and Luca Wehrstedt and Madian Khabsa and Manav Avalani and Manish Bhatt and Martynas Mankus and Matan Hasson and Matthew Lennie and Matthias Reso and Maxim Groshev and Maxim Naumov and Maya Lathi and Meghan Keneally and Miao Liu and Michael L. Seltzer and Michal Valko and Michelle Restrepo and Mihir Patel and Mik Vyatskov and Mikayel Samvelyan and Mike Clark and Mike Macey and Mike Wang and Miquel Jubert Hermoso and Mo Metanat and Mohammad Rastegari and Munish Bansal and Nandhini Santhanam and Natascha Parks and Natasha White and Navyata Bawa and Nayan Singhal and Nick Egebo and Nicolas Usunier and Nikhil Mehta and Nikolay Pavlovich Laptev and Ning Dong and Norman Cheng and Oleg Chernoguz and Olivia Hart and Omkar Salpekar and Ozlem Kalinli and Parkin Kent and Parth Parekh and Paul Saab and Pavan Balaji and Pedro Rittner and Philip Bontrager and Pierre Roux and Piotr Dollar and Polina Zvyagina and Prashant Ratanchandani and Pritish Yuvraj and Qian Liang and Rachad Alao and Rachel Rodriguez and Rafi Ayub and Raghotham Murthy and Raghu Nayani and Rahul Mitra and Rangaprabhu Parthasarathy and Raymond Li and Rebekkah Hogan and Robin Battey and Rocky Wang and Russ Howes and Ruty Rinott and Sachin Mehta and Sachin Siby and Sai Jayesh Bondu and Samyak Datta and Sara Chugh and Sara Hunt and Sargun Dhillon and Sasha Sidorov and Satadru Pan and Saurabh Mahajan and Saurabh Verma and Seiji Yamamoto and Sharadh Ramaswamy and Shaun Lindsay and Shaun Lindsay and Sheng Feng and Shenghao Lin and Shengxin Cindy Zha and Shishir Patil and Shiva Shankar and Shuqiang Zhang and Shuqiang Zhang and Sinong Wang and Sneha Agarwal and Soji Sajuyigbe and Soumith Chintala and Stephanie Max and Stephen Chen and Steve Kehoe and Steve Satterfield and Sudarshan Govindaprasad and Sumit Gupta and Summer Deng and Sungmin Cho and Sunny Virk and Suraj Subramanian and Sy Choudhury and Sydney Goldman and Tal Remez and Tamar Glaser and Tamara Best and Thilo Koehler and Thomas Robinson and Tianhe Li and Tianjun Zhang and Tim Matthews and Timothy Chou and Tzook Shaked and Varun Vontimitta and Victoria Ajayi and Victoria Montanez and Vijai Mohan and Vinay Satish Kumar and Vishal Mangla and Vlad Ionescu and Vlad Poenaru and Vlad Tiberiu Mihailescu and Vladimir Ivanov and Wei Li and Wenchen Wang and Wenwen Jiang and Wes Bouaziz and Will Constable and Xiaocheng Tang and Xiaojian Wu and Xiaolan Wang and Xilun Wu and Xinbo Gao and Yaniv Kleinman and Yanjun Chen and Ye Hu and Ye Jia and Ye Qi and Yenda Li and Yilin Zhang and Ying Zhang and Yossi Adi and Youngjin Nam and Yu and Wang and Yu Zhao and Yuchen Hao and Yundi Qian and Yunlu Li and Yuzi He and Zach Rait and Zachary DeVito and Zef Rosnbrick and Zhaoduo Wen and Zhenyu Yang and Zhiwei Zhao and Zhiyu Ma},
      year={2024},
      eprint={2407.21783},
      archivePrefix={arXiv},
      primaryClass={cs.AI},
      url={https://arxiv.org/abs/2407.21783}, 
}

@online{economic_toll,
  title = {The Economic Toll of Fake Reviews: Market Data and Prevention Strategies},
  author = {Max Chekalov},
  url = {https://www.99firms.com/blog/the-economic-toll-of-fake-reviews/?utm_source=chatgpt.com},
  year = {2024},
}

@misc{openai2025gpt41,
  author       = {OpenAI},
  title        = {Introducing GPT-4.1 in the API},
  year         = {2025},
  howpublished = {\url{https://openai.com/index/gpt-4-1/}},
  note         = {Accessed: 2025-05-22}
}

@inproceedings{papineni-etal-2002-bleu,
  title     = "{B}leu: a Method for Automatic Evaluation of Machine Translation",
  author    = "Papineni, Kishore and Roukos, Salim and Ward, Todd and Zhu, Wei-Jing",
  booktitle = "Proceedings of the 40th Annual Meeting of the Association for Computational Linguistics",
  year      = "2002",
  pages     = "311--318",
  address   = "Philadelphia, Pennsylvania, USA",
  publisher = "Association for Computational Linguistics",
  url       = "https://aclanthology.org/P02-1040/",
  doi       = "10.3115/1073083.1073135"
}

@misc{he2015delvingdeeprectifierssurpassing,
      title={Delving Deep into Rectifiers: Surpassing Human-Level Performance on ImageNet Classification}, 
      author={Kaiming He and Xiangyu Zhang and Shaoqing Ren and Jian Sun},
      year={2015},
      eprint={1502.01852},
      archivePrefix={arXiv},
      primaryClass={cs.CV},
      url={https://arxiv.org/abs/1502.01852}, 
}

@article{kingma2014adam,
  title       = {Adam: A Method for Stochastic Optimization},
  author      = {Kingma, Diederik P. and Ba, Jimmy},
  journal     = {arXiv preprint arXiv:1412.6980},
  year        = {2014},
  url         = {https://arxiv.org/abs/1412.6980}
}

@misc{xu2025nucleardeployedanalyzingcatastrophic,
      title={Nuclear Deployed: Analyzing Catastrophic Risks in Decision-making of Autonomous LLM Agents}, 
      author={Rongwu Xu and Xiaojian Li and Shuo Chen and Wei Xu},
      year={2025},
      eprint={2502.11355},
      archivePrefix={arXiv},
      primaryClass={cs.CL},
      url={https://arxiv.org/abs/2502.11355}, 
}

@article{gu2024survey,
  title={A survey on llm-as-a-judge},
  author={Gu, Jiawei and Jiang, Xuhui and Shi, Zhichao and Tan, Hexiang and Zhai, Xuehao and Xu, Chengjin and Li, Wei and Shen, Yinghan and Ma, Shengjie and Liu, Honghao and others},
  journal={arXiv preprint arXiv:2411.15594},
  year={2024}
}
